%% file: 0-main.tex
\documentclass[letterpaper]{article} % DO NOT CHANGE THIS
\usepackage{aaai24}  % DO NOT CHANGE THIS
\usepackage{times}  % DO NOT CHANGE THIS
\usepackage{helvet}  % DO NOT CHANGE THIS
\usepackage{courier}  % DO NOT CHANGE THIS
\usepackage[hyphens]{url}  % DO NOT CHANGE THIS
\usepackage{graphicx} % DO NOT CHANGE THIS
\usepackage{natbib}  % DO NOT CHANGE THIS AND DO NOT ADD ANY OPTIONS TO IT
\usepackage{caption} % DO NOT CHANGE THIS AND DO NOT ADD ANY OPTIONS TO IT
\usepackage{comment}
\usepackage{amsmath}
\usepackage{amssymb}

\usepackage{booktabs}
\usepackage{url}
\usepackage{enumitem}
\usepackage{caption}
\usepackage{xspace}
\usepackage{xcolor}
\usepackage{subcaption}
\usepackage{amssymb}
\usepackage{algorithm}
\usepackage{algorithmic}

\usepackage{enumitem}
\usepackage{booktabs}
\usepackage{makecell}
\usepackage{ctable}
\usepackage{xcolor}
\usepackage{amsfonts}
\usepackage{multirow}
\usepackage{comment}
\usepackage{comment}
\usepackage{newfloat}
\usepackage{listings}
\DeclareCaptionStyle{ruled}{labelfont=normalfont,labelsep=colon,strut=off} % DO NOT CHANGE THIS
\floatstyle{ruled}
\newfloat{listing}{tb}{lst}{}
\floatname{listing}{Listing}
\title{Toward Robustness in Multi-label Classification: \\A Data Augmentation Strategy against Imbalance and Noise}
\author{
Hwanjun Song\textsuperscript{\rm 1},
Minseok Kim\textsuperscript{\rm 2},
Jae-Gil Lee\textsuperscript{\rm 1}}
\affiliations{
    \textsuperscript{\rm 1}KAIST,~~\textsuperscript{\rm 2}Amazon\\
    {\{songhwanjun, jaegil\}}@kaist.ac.kr, kminseok@amazon.com
}

\usepackage{bibentry}

\newcolumntype{L}[1]{>{\raggedright\let\newline\\\arraybackslash\hspace{0pt}}m{#1}}
\newcolumntype{X}[1]{>{\centering\let\newline\\\arraybackslash\hspace{0pt}}p{#1}}
\newcolumntype{Y}[1]{>{\raggedleft\let\newline\\\arraybackslash\hspace{0pt}}m{#1}}

\definecolor{Gray}{gray}{0.90}
\newcommand{\algname}{{BalanceMix}} % Federated learning with Reliable Neighbors

\newcommand{\INDSTATE}[1][1]{\STATE\hspace{#1\algorithmicindent}}
\definecolor{Gray}{gray}{0.90}

\begin{document}

\maketitle

\begin{abstract}
Multi-label classification poses challenges due to imbalanced and noisy labels in training data. We propose a unified data augmentation method, named BalanceMix, to address these challenges. Our approach includes two samplers for imbalanced labels, generating minority-augmented instances with high diversity. It also refines multi-labels at the label-wise granularity, categorizing noisy labels as clean, re-labeled, or ambiguous for robust optimization. Extensive experiments on three benchmark datasets demonstrate that BalanceMix outperforms existing state-of-the-art methods. We release the code at {\url{https://github.com/DISL-Lab/BalanceMix}}.
\end{abstract}

\input{1-introduction}
\input{2-relatedwork}
\input{3-method}

\input{4-experiment}
\input{5-conclusion}

\input{0-main.bbl}
\bibliography{aaai24}

\clearpage
\input{6-appendix}

\end{document}

%% file: 1-introduction.tex
\section{Introduction}
\label{sec:introduction}

{The issue of data-label quality emerges as a major concern in the practical use of deep learning, potentially resulting in catastrophic failures when deploying models in real-world test scenarios\,\cite{whang2021data}. This concern is magnified in multi-label classification, where instances can be associated with multiple labels simultaneously. In this context, AI system robustness is at risk due to diverse types of data-label issues, although the task can reflect the complex relationships present in real-world data\,\cite{bello2021data}.}

The presence of \emph{class imbalance} occurs when a few majority classes occupy most of the positive labels, and \emph{positive-negative imbalance} arises due to instances typically having fewer positive labels but numerous negative labels. Such imbalanced labels can dominate the optimization process and lead to underemphasizing the gradients from minority classes or positive labels.
{Additionally, the presence of \emph{noisy labels} stems from the costly and time-consuming nature of meticulous annotation\,\cite{song2022learning}}. Labels can be corrupted by adversaries or system failures\,\cite{zhang2020adversarial}. Notably, instances have both clean and incorrect labels, therefore resulting in diverse cases of noisy labels. 

\begin{figure}[t!]
\begin{center}
\includegraphics[width=7.7cm]{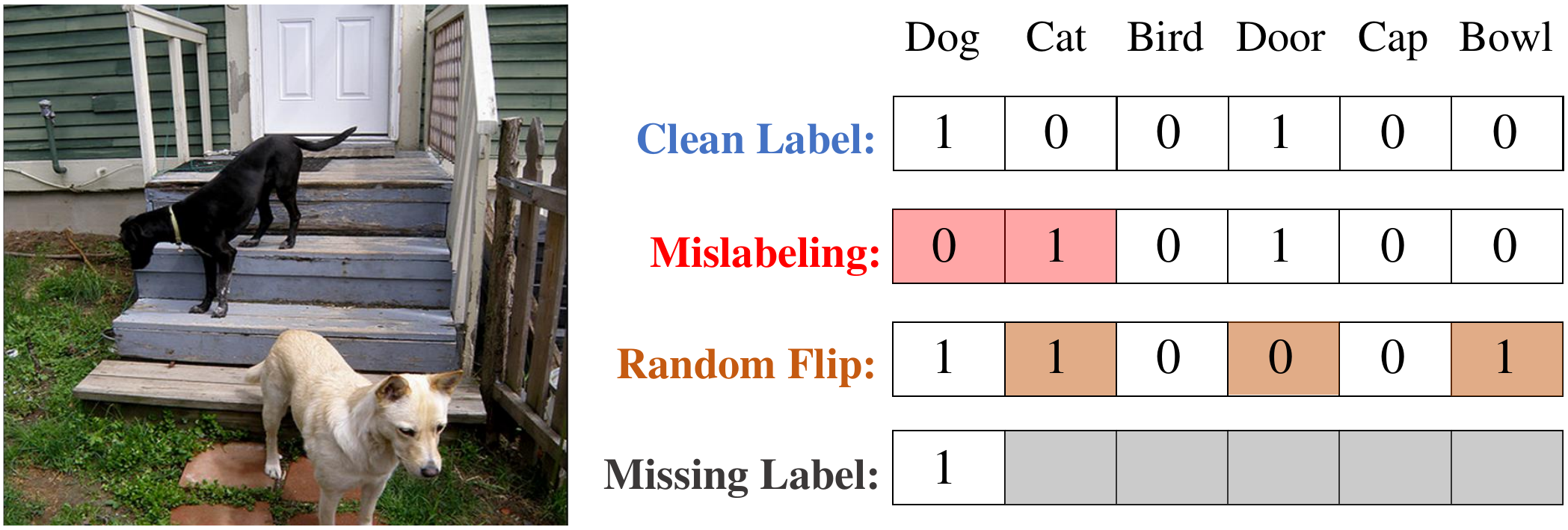}
\end{center}
\vspace*{-0.35cm}
\caption{Examples of noisy labels in multi-label classification. $1$ and $0$ indicate positive and negative labels, symbolizing the existence of an object class.}
\label{fig:label_problem}
\vspace*{-0.6cm}
\end{figure}

Three distinct types of noisy labels arise in multi-label classification, as illustrated in Fig.~\ref{fig:label_problem}: \emph{mislabeling}, where a visible object in the image is labeled incorrectly by a human or machine annotator, such as a dog being labeled as a cat; \emph{random flipping}, where labels are randomly flipped by an adversary regardless of the presence of other class objects, such as negative labels for a cat and a bowl being flipped independently to positive labels; and \emph{(partially) missing labels}, where even humans cannot find all applicable class labels for each image, and it is more difficult to detect their absence than to detect their presence\,\cite{cole2021multi}. %As a result, missing labels are typically treated as noisy negative labels in training, as indicated by the gray labels in Fig.\ \ref{fig:label_problem}.

\begin{figure*}[t!]
\begin{center}
\includegraphics[width=17.4cm]{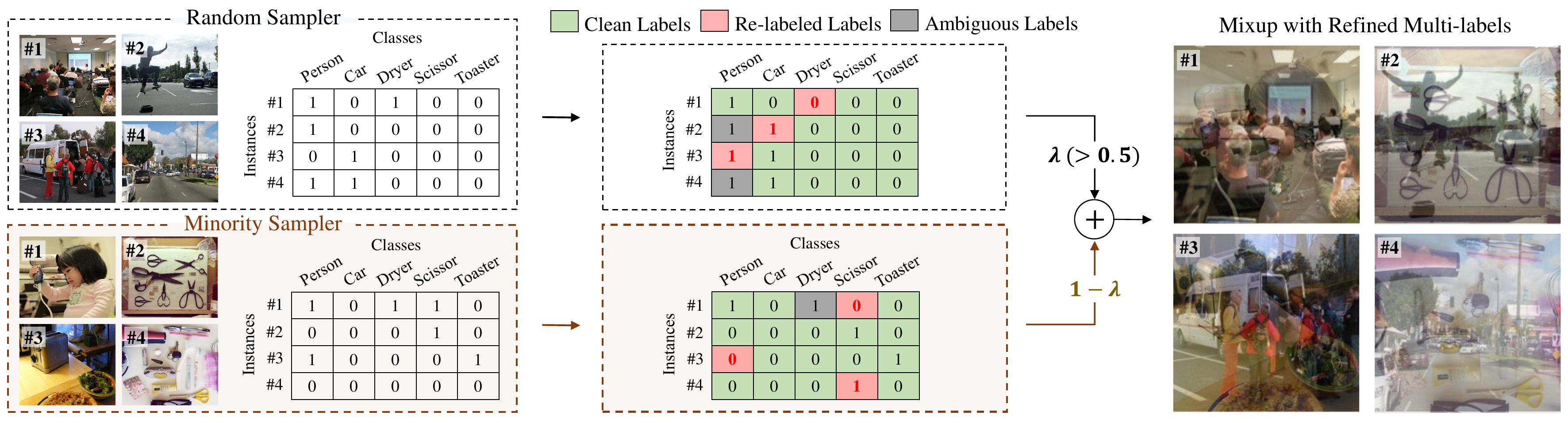}
\end{center}
\vspace*{-0.2cm}
\hspace*{0.05cm}{\small (a) Minority Sampling (w.\ Noisy Multi-labels).} \hspace*{0.5cm}{\small (b) Fine-grained Label-wise Management.} \hspace*{1.35cm}{\small (c) Mixing Augmentation.}
% \hspace*{0.05cm}{\small (a) Sampled Instances with Noisy Multi-labels.} \hspace*{0.4cm}{\small (b) Fine-grained Label-wise Management.} \hspace*{1.85cm}{\small (c) Mixing Augmentation.}
\vspace*{-0.2cm}
\caption{{Overview of BalanceMix.} Here with MS-COCO, ``person'' and ``car'' are the most frequently observed\,(majority) classes, whereas ``hair dryer,'' ``scissor,'' and ``toaster'' are the least frequently observed\,(minority) classes in the training data.}
\vspace*{-0.4cm}
\label{fig:overview}
\end{figure*}

{
Ensuring the robustness of AI systems calls for a \emph{holistic} approach that effectively operates within the following settings: clean, noisy, missing, and imbalanced labels at the same time. However, this task is non-trivial given that minority and noisy labels have similar behavior in learning, e.g., larger gradients, making the task even more complicated and challenging.
As a result,} prior studies have addressed these two problems \emph{separately} in different setups, assuming either clean or well-balanced training data---i.e.,
imbalanced clean labels\,\cite{lin2017focal, ben2020asymmetric, du2023superdisco} and well-balanced noisy labels\,\cite{zhao2021evaluating, ferreira2021explainable, li2022estimating, wei2023fine}.

{We address this challenge using a new data augmentation method,  \textbf{BalanceMix}, without complex data preprocessing and architecture change.}
%we propose a simple yet effective data augmentation method named \textbf{BalanceMix} for balanced and robust optimization. 
First, for imbalanced multi-labels, we maintain an additional batch sampler called a \emph{minority sampler}, which samples the instances containing minority labels with high probability, as illustrated in Fig.~\ref{fig:overview}(a). To counter the limited diversity in oversampling, we interpolate the instances sampled from the minority sampler with those sampled from a random sampler using the Mixup\,\cite{zhang2017mixup} augmentation. 
By mixing with a higher weight to the random instances, the sparse context of the oversampled instances literally \emph{percolates} through the majority of training data without losing diversity. 
Minority sampling in Fig.~\ref{fig:overview}(a) followed by the Mixup augmentation in Fig.~\ref{fig:overview}(c) is called \emph{minority-augmented mixing}.

Then, for noisy multi-labels, we incorporate \emph{fine-grained label-wise management} to feed high-quality multi-labels into the augmentation process. Unlike existing robust learning methods such as Co-teaching\,\cite{han2018co} which consider each instance as a candidate for selection or correction, we should move to a finer granularity and consider \emph{each label} as a candidate.
As illustrated in Fig.~\ref{fig:overview}(b), the label-wise management step categorizes the entire set of noisy labels into three subsets: \emph{clean} labels which are expected to be correct with high probability;  \emph{re-labeled} labels whose flipping is corrected with high confidence; and  \emph{ambiguous} labels which need to be downgraded in optimization. 
Putting our solutions for imbalanced and noisy labels together, % i.e., by mixing up the instances individually chosen from the two samplers with refined multi-labels, 
\algname{} % featured with \textrm{minority-augmented mixing} and \textrm{fine-grained label-wise management}
is completed, as illustrated in Fig.~\ref{fig:overview}. % In the presence of imbalanced and noisy

%While \algname{} utilizes Mixup, the novelty of this work should not underestimate simply due to the use of a familiar technique. The true technical innovation lies in successfully integrating this well-known technique into a practical and challenging problem, making our approach unique and noteworthy (see Section \ref{sec:methodology} for details).
{
Our technical innovations are successfully incorporated into the well-established techniques of oversampling and Mixup, enabling easy integration into the existing training pipeline. % like ML-Decoder\,\cite{ridnik2023ml}.
Our contributions are threefold: (1) \algname{} serves as a versatile data augmentation technique, demonstrating reliable performance across clean, noisy, missing, and imbalanced labels.} (2) \algname{} avoids overfitting to minority classes and incorrect labels thanks to minority-augmented mixing with fine-grained label management. %(3) \algname{} outperforms existing prior arts in each setting of three different types of label noise with the co-existence of severe class imbalance. 
(3) \algname{} outperforms existing prior arts and reaches 91.7mAP on the MS-COCO data, which is the state-of-the-art performance with the ResNet backbone. 

% The novelty of this work should not be underestimated due to the use of a familiar technique

%{We believe that our work is a promising approach to combining imbalanced and noisy label problems with multi-label classification research.}

\begin{comment}
%
\begin{itemize}[leftmargin=10pt]
\setlength\itemsep{0em}
% \item To our knowledge, this is the first work to address both imbalanced and noisy labels {\color{purple}from a data augmentation perspective.}
\item \algname{} is a data augmentation technique that can be easily integrated into various end-to-end deep learning algorithm for multi-label classification.
\item \algname{} avoids the overfitting to minority classes and incorrect labels thanks to minority-augmented mixing with fine-grained label management.
\item \algname{} outperforms existing state-of-the-art methods in each setting of three different types of label noise with the co-existence of severe class imbalance.
\item \algname{} reaches 91.7mAP on the MS-COCO data, which is the state-of-the-art performance with the ResNet backbone in multi-label classification.
%\algname{} sets a new record of 91.7mAP in multi-label classification on the MS-COCO data.
\end{itemize}
\end{comment}

%% file: 2-relatedwork.tex
\section{Related Work}
\label{sec:preliminary}

\paragraph{Multi-label with Imbalance.} 
One of the main trends in this field is solving long-tail class imbalance and positive-negative label imbalance. There have been classical resampling approaches\,\cite{wang2014resampling, loyola2016study} for imbalance, but they are mostly designed for a single-label setup. A common solution for class imbalance with multi-labels is the focal loss \cite{lin2017focal}, which down-weights the loss value of each label gradually as a model's prediction confidence increases, highlighting difficult-to-learn minority class labels; however, it can lead to overfitting to incorrect labels. % since they are also difficult-to-learn. Another attempt has been made to address the positive-negative imbalance while resolving its weakness.
The asymmetric focal loss\,(ASL)\,\citep{ben2020asymmetric} modifies the focal loss to operate differently on positive and negative labels for the imbalance. \citep{yuan2023balanced} proposed a balance masking strategy using a graph-based approach.

% and adopts probability shifting to remove possibly incorrect negative labels. %However, 
%based on our analysis in Section X, 
%However, it trashes a significant amount of available labels in the setup of random flipping and missing labels.

% importance controal -> risk of emphasizing noisy instances..

\paragraph{Multi-label with (Partially) Missing Labels.} 
Annotation in the multi-label setup becomes harder as the number of classes increases. Subsequently, the need to handle missing labels has recently gained a lot of attention. A simple solution is regarding all the missing labels as negative labels \cite{wang2014binary}, but it leads to overfitting to incorrect negative ones. 
There have been several studies with deep neural networks\,(DNNs). \citep{durand2019learning} adopted curriculum learning for pseudo-labeling based on model predictions. \citep{huynh2020interactive} used the inferred dependencies among labels and images to prevent overfitting. Recently, \citep{cole2021multi} and \citep{kim2022large, kim2023bridging} addressed the hardest version, where only a single positive label is provided for each instance. They proposed multiple solutions including label smoothing, expected positive regularization, and label selection and correction. However, the imbalance problem is overlooked, and all the labels are simply assumed to be clean.

% only consider negative cases, assmuping positive labels are all cleans..

\paragraph{Classification with Noisy Labels.} For \emph{single}-label classification, learning with noisy labels has established multiple directions. %, such as sample selection. , loss correction, re-labeling, and semi-supervised learning\,\cite{song2022learning}. 
Most approaches are based on the memorization effect of DNNs, in which simple and generalized patterns are prone to be learned before the overfitting to noisy patterns\,\cite{arpit2017closer}. 
More specifically, instances with small losses or consistent predictions are treated as clean instances, as in Co-teaching\,\cite{han2018co}, O2U-Net\,\cite{huang2019o2u}, and CNLCU\,\cite{xia2021sample}; instances are re-labeled based on a model's predictions for label correction, as in SELFIE\,\cite{song2019selfie} and SEAL\,\cite{chen2021beyond}. 
A considerable effort has also been made to use semi-supervised learning, as in DivideMix\,\cite{li2020dividemix} and PES\,\cite{bai2021understanding}. 
In addition, a few studies have addressed class imbalance in the noisy single-label setup\,\cite{wei2021robust, ding2022multi}, but they cannot be immediately applied to the multi-label setup owing to their inability to handle the various types of label noise caused by the nature of having both clean and incorrect labels in one instance.

For \emph{multi}-label classification with noisy labels, there has yet to be studied actively owing to the inherent complexity including diverse types of label noise and imbalance. CbMLC\,\cite{zhao2021evaluating} addresses label noise by proposing a context-based classifier, but its architecture is confined to graph neural networks and requires large pre-trained word embeddings. A method by Hu et al.\cite{hu2018multi} utilizes a teacher-student network with feature transformation. SELF-ML\,\cite{ferreira2021explainable} re-labels an incorrect label using a combination of clean labels, but it works only when multi-labels can be defined as attributes associated with each other. %, e.g., attributes of clothing. Most recently, 
ASL\,\cite{ben2020asymmetric} solves the problem of mislabeling by shifting the prediction probability of low-confidence negative labels, making their losses close to zero in optimization. T-estimator\,\cite{li2022estimating} solves the estimation problem of the noise transition matrices in the multi-label setting.

%\smallskip\smallskip
%\noindent\textbf{Novelty.} Differently from past works that only focus on a specific objective, i.e., either imbalance or label noise, we propose a simple yet comprehensive approach from the perspective of data augmentation, which allows the adoption in other multi-label classification methods.

\paragraph{Oversampling with Mixup.} 
Prior studies have applied Mixup to address class imbalance \cite{guo2021long, wu2020distribution, galdran2021balanced, park2022majority}. Yet, they mainly focus on \emph{single}-label classification, overlooking positive-negative imbalances and noisy labels.
We propose the first approach that uses predictive confidence to dynamically adjust the degree of oversampling for both types of imbalance while employing label-wise management for noisy labels. %Our approach is a unified data augmentation process that is compatible with existing training pipelines, making it easily applicable to various real-world scenarios.

%% file: 3-method.tex
\section{Problem Definition}
\label{sec:problem}

A multi-label multi-class classification problem requires training data $\mathcal{D}$, a set of two random variables ($\mathbf{x}$, $\mathbf{y}$) which consists of an instance ($d$-dimensional feature) $\mathbf{x}$ $\in \mathcal{X}$\,$(\subset \mathbb{R}^{d})$ and its multi-label $\mathbf{y}$ $\in \{0, 1\}^{K}$, where $K$ is the number of applicable classes. However, in the presence of label noise, the noisy multi-label $\tilde{\mathbf{y}}\in \{0, 1\}^{K}$ possibly contains incorrect labels originated from mislabeling, random flipping, and missing labels; that is, a noisy label $\tilde{y_k} \in \tilde{\mathbf{y}}$ may not be equal to the true label ${y_k} \in {\mathbf{y}}$. Thus, let $\tilde{\mathcal{D}}=\{(\mathbf{x}_n, \tilde{\mathbf{y}}_n)\}_{n=1}^{N}$ be the noisy training data of size $N$. % in multi-label classification with noisy labels. 

% JGL: don't use "imperfect" labels -> "noisy" labels, use "imbalanced" explicitly if necessary.
\paragraph{Label Noise Modeling.} We define three types of label noise. From the statistical point of view, (1) \emph{mislabeling} is defined as \emph{class-dependent} label noise, where a class object in the image is incorrectly labeled as another class object that may not be visible. The ratio of a class $c_1$ being mislabeled as $c_2$ is formulated by $\rho_{c_1 \rightarrow c_2} \!\!=\!\! p(\tilde{y}_{c_1}\!\!=\!\!0, \tilde{y}_{c_2}\!\!=\!\!1 | y_{c_1}\!\!=\!\!1, y_{c_2}\!\!=\!\!0)$. In contrast, (2) \emph{random flipping} is \emph{class-independent} label noise, where the presence\,(or absence) of a class $c$ is randomly flipped with a probability of $\rho_{c}=p(\tilde{y}_{c}=1|y_c=0)=p(\tilde{y}_{c}=0|y_c=1)$, which is independent of the presence of other classes. This scenario can be caused by an adversary's attack or a system failure. Last, (3) \emph{missing labels} from partial labeling can be considered as a type of label noise, where all missing labels are treated as negative ones.%  part of positive labels are omitted;

\paragraph{Optimization.} To deal with multi-labels in optimization, the most widely-used approach is solving $K$ binary classification problems using the binary cross-entropy (BCE) loss. Given a DNN parameterized by $\Theta$, the DNN is updated via stochastic gradient descent to minimize the expected BCE loss on the mini-batch ${B}\subset \tilde{\mathcal{D}}$,
\vspace*{-0.15cm}
\begin{equation}
\small
\begin{split}
\mathcal{L}(B; \Theta) &= \frac{1}{|B|} \sum_{(\mathbf{x}, \tilde{\mathbf{y}})\in B}\sum_{k=1}^{K} {\rm BCE}(f_{(\mathbf{x}, \tilde{y}_k)}), ~{\rm where} \\
\label{eq:standard_update}
{\rm BCE}(f_{(\mathbf{x}, \tilde{y}_k)}) &= -\tilde{y}_k \cdot {\rm log}(f_{(\mathbf{x},\tilde{y}_k)}) - ( 1\!-\!\tilde{y}_k) \cdot {\rm log}(1\!-\!f_{(\mathbf{x},\tilde{y}_k)}).
\end{split}
\raisetag{30pt}
\end{equation}
Given the instance $\mathbf{x}$, $f_{(\mathbf{x},\tilde{y}_k)}$ and $1\!-\!f_{(\mathbf{x},\tilde{y}_k)}$ are the confidence in presence and absence, respectively, for the $k$-th class by the model $\Theta$. \algname{} is built on top of this standard optimization pipeline for multi-label classification. 

\section{Methodology: BalanceMix}
\label{sec:methodology}

Our primary idea is to generate minority-augmented instances and their reliable multi-labels through \emph{data augmentation}. We now detail the two main components, which achieve balanced and robust optimization by \emph{minority-augmented mixing} and \emph{label-wise management}. The pseudocode of \algname{} is provided in {Appendix A}. 
%These two components are described consecutively for smooth presentation, but label-wise management (Fig.~\ref{fig:overview}(b)) is done in between minority sampling (Fig.~\ref{fig:overview}(a)) and mixing augmentation (Fig.~\ref{fig:overview}(c)), two sub-components of minority-augmented mixing.

\subsection{Minority-augmented Mixing}
\label{sec:minority_mixing}

To relieve the class imbalance problem, prior studies either oversample the minority class labels or adjust their loss values\,\cite{tarekegn2021review}. These methods are intuitive but rather intensify the overfitting problem since they rely on a few minority instances with limited diversity\,\cite{guo2021long}.  
{On the other hand, we leverage random instances to increase the diversity of minority instances by \emph{separately} maintaining two samplers in Fig.~\ref{fig:overview}.}

%we generate minority-augmented instances with high diversity by \emph{separately} maintaining random and minority samplers in Fig.~\ref{fig:overview}. 
%The random sampler draws instances at random from training data, and we concentrate on the \emph{minority sampler}.
%There have been a few studies based on augmentation\,\cite{shamsolmoali2021imbalanced, park2022majority}, but they deal with the \emph{single}-label classification. % where each instance has only a single class object. 

\begin{figure}[t!]
\vspace*{-0.05cm}
\begin{center}
\includegraphics[width=8.4cm]{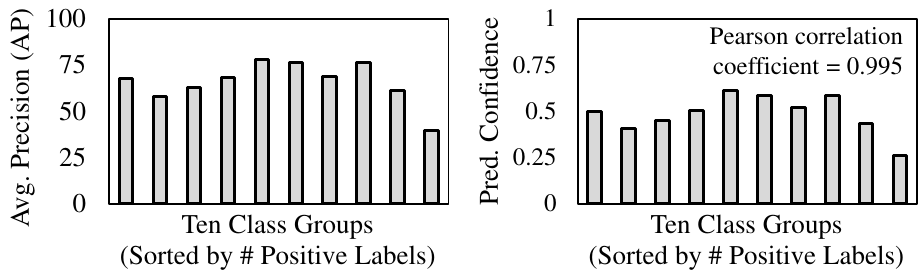}
\end{center}
\vspace*{-0.2cm}
\hspace*{0.7cm} {\small (a) Prediction Confidence.} \hspace*{0.5cm} {\small (b) Average Precision\,(AP).} \!\!\!
\vspace*{-0.2cm}
\caption{AP and Prediction confidence in COCO using the BCE loss at the 40$\%$ of training epochs, where 80 classes are partitioned into ten groups in the descending order of positive label frequency. The Pearson correlation coefficient is computed between ten class groups.}
\label{fig:sampler_motivation}
\vspace*{-0.47cm}
\end{figure}

\paragraph{Confidence-based Minority Sampling.} %The first question here is which instances should be oversampled to improve the average precision\,(AP) of the minority labels. However, 
{
Prior oversampling methods that rely on the frequency of positive labels face two key limitations. First, this frequency alone does not identify the minority multi-labels with low AP values; as illustrated in Fig.~\ref{fig:sampler_motivation}(a), the class group with few positive labels does not always have lower AP values due to the complexity of the two types of imbalance in the multi-label setup. Second, there is a risk of overfitting because of sticking to the same oversampling policy during the entire training period.
}

To address these limitations, we first propose to employ the prediction confidence $f_{(\mathbf{x}, \tilde{y}_k)}$, which exhibits a strong correlation with the AP, as shown in Fig.~\ref{fig:sampler_motivation}(b). We opt to oversample the instances with low prediction confidence in their multi-labels, as they are expected to contribute the most significant increase in the AP. Initially, We define two confidence scores for a \emph{specific} class $k$,
%Oversampling based solely on positive label frequency is suboptimal as it disregards the complexity of the two types of imbalance, i.e., class- and positive-negative imbalance that co-occur. Upon comparing Fig.~\ref{fig:sampler_motivation}(a) and Fig.~\ref{fig:sampler_motivation}(b), the class group with many positive labels does not always have higher AP values. 
%On the other hand, prediction confidence $f_{(\mathbf{x}, \tilde{y}_k)}$ exhibits a strong correlation with the AP. Therefore, we oversample the instances which have low prediction confidence in their multi-labels since they are expected to increase the AP the most. We first define two confidence scores for a \emph{specific} class $k$,
\begin{equation}
\small
\begin{split}
{\rm P}(k) =  \frac{1}{|{\mathcal{P}_k}|}\!\sum_{(\mathbf{x}, \tilde{\mathbf{y}}) \in {\mathcal{P}_k}}\!\!\!\!\! f_{(\mathbf{x},\tilde{y}_k)},~~ {\rm A}(k) =  \frac{1}{|{\mathcal{A}_k}|}\!\sum_{(\mathbf{x}, \tilde{\mathbf{y}}) \in {\mathcal{A}_k}}\!\!\!\! (1\!-\!f_{(\mathbf{x},\tilde{y}_k)}),\\
{\rm s.t.} ~ \mathcal{P}_k=\{(\mathbf{x}, \tilde{\mathbf{y}}) \in \tilde{\mathcal{D}}: \tilde{y}_k\!=\! 1\}, ~ \mathcal{A}_k=\{(\mathbf{x}, \tilde{\mathbf{y}}) \in \tilde{\mathcal{D}}: \tilde{y}_k\! = \!0\},
\end{split}
\label{eq:conf_score_class}
\raisetag{25pt}
\end{equation}
which are the expected prediction confidences, respectively, for the presence\,(P) and absence\,(A) of the $k$-th class. % Even with label noise, the trend of confidence scores is statistically similar to those in clean data due to the dominance of clean labels in training data.
Next, the confidence score of an instance $(\mathbf{x}, \tilde{\mathbf{y}})$ is defined by aggregating Eq.\ \eqref{eq:conf_score_class} for \emph{all} the classes,
\begin{equation}
\small
\text{Score}(\mathbf{x}, \tilde{\mathbf{y}}) = \sum_{k=1}^{K} \mathbf{1}_{[\tilde{y}_k=1]}{\rm P}(k) + \mathbf{1}_{[\tilde{y}_k=0]}{\rm A}(k).
\end{equation}
Then, the sampling probability of $(\mathbf{x}, \tilde{\mathbf{y}})$ is formulated as
\begin{equation}
p_{\text{sampling}}\big((\mathbf{x}, \tilde{\mathbf{y}}); \tilde{\mathcal{D}}\big) = \frac{1 / \text{Score}(\mathbf{x}, \tilde{\mathbf{y}})}{\sum_{(\mathbf{x}^{\prime}, \tilde{\mathbf{y}}^{\prime})\in\tilde{\mathcal{D}}} 1 / {\rm Score}(\mathbf{x}^{\prime}, \tilde{\mathbf{y}}^{\prime})}.
\label{eq:weights}
\end{equation}

\begin{comment}
    
\setlength{\columnsep}{4.5pt} % left margin
\begin{wrapfigure}{r}{0.5\linewidth}
\vspace{-1.0em}
\hspace{2cm}
\includegraphics[width=1\linewidth]{iccv2023AuthorKit/figures/methodology/sampling.pdf}
\vspace{-1.5em}
\caption{Sampling probability.}
\label{fig:prob}
\vspace{0em}
\end{wrapfigure}
\end{comment}

By doing so, we consider {positive-negative imbalance together with class imbalance} by relying on the prediction confidence, which marks a significant difference from existing methods\,\cite{galdran2021balanced, park2022majority} that considers only the class imbalance of positive labels. % for the single-label setup. 
Further, our {confidence-based} sampling dynamically adjusts the degree of oversampling over a training period, {mitigating the risk of overfitting.} Minority instances are initially oversampled with a high probability, but the degree of oversampling gradually decreases as the imbalance problem gets resolved (see Figure \ref{fig:sampling_analysis} in Appendix B).

%, as shown in the red line of Fig.\ \ref{fig:prob}.
\begin{comment}
Therefore, the instances with minority labels are oversampled since they exhibit lower prediction confidence in presence (for positive labels) and absence (for negative labels).

\setlength{\columnsep}{4.5pt} % left margin
\begin{wrapfigure}{r}{0.5\linewidth}
\vspace{-1.0em}
\hspace{2cm}
\includegraphics[width=1\linewidth]{iccv2023AuthorKit/figures/methodology/sampling.pdf}
\vspace{-1.5em}
\caption{Sampling probability by the minority sampler.}
\label{fig:prob}
\vspace{0em}
\end{wrapfigure}Balanced-mixup considers only the class imbalance of positive labels for the single-label setup. However, we consider the \emph{positive-negative imbalance together with the class imbalance} by relying on the prediction confidence (see Eq.\ (2)). In particular, our \emph{confidence-based} sampling makes a big difference in that it adjusts the degree of oversampling over training, as shown in Fig.\ \ref{fig:prob}. Minority instances are initially oversampled with a high probability, but the degree of oversampling gradually decreases as the imbalance problem gets resolved. In contrast, frequency-based sampling, such as Balanced-mixup, can lead to overfitting because it sticks to the same oversampling policy during the entire period.
\end{comment}

\paragraph{Mixing Augmentation.} 
To mix the instances from the two samplers. We adopt the Mixup \cite{zhang2017mixup} augmentation because it can mix two instances even when multi-labels are assigned to them. Let $(\mathbf{x}_{R}, \tilde{\mathbf{y}}_{R})$ and $(\mathbf{x}_{M}, \tilde{\mathbf{y}}_{M})$ be the instances sampled from the random and minority samplers, respectively. The minority-augmented instance is generated by their interpolation, 
\begin{equation}
\begin{gathered}
\mathbf{x}^{mix}\!=\! \lambda\mathbf{x}_{R} \!+\! (1\!-\!\lambda) \mathbf{x}_{M}, ~\tilde{\mathbf{y}}^{mix}\!= \!\lambda \tilde{\mathbf{y}}_{R} \!+\! (1\!-\!\lambda) \tilde{\mathbf{y}}_{M}\!\!\\
{\rm where} ~~ \lambda = {\rm max}(\lambda^{\prime} , 1-\lambda^{\prime} ),
\end{gathered}
\label{eq:mixup}
\end{equation}
and $\lambda^{\prime}  \in [0,1] \sim {\rm Beta}(\alpha, \alpha)$. 
By the second row of Eq.~\eqref{eq:mixup}, $\lambda$ becomes greater than or equal to $0.5$; thus, the instance of the random sampler amplifies diversity, while that of the minority sampler adds the context of minority classes. \emph{Mixing one random instance and one controlled\,(minor) instance}, instead of mixing two random instances, is a simple yet effective strategy, as shown in the evaluation.
% Random samplers increase the diversity of instances, but lead to imbalanced optimization in the presence of imbalanced labels. In contrast, the random sampler mitigates the imbalance problem by oversampling instances with low confidence, but overfitting occurs due to limited diversity. To maintain high diversity in balanced optimization, we force the instances from the random sampler dominates in mixing by the second row of Eq.~\eqref{eq:mixup}; the $\lambda$ becomes greater than or equal to $0.5$. Therefore, the minority sampler adds the sparse context of minority labels without losing their diversity.

% JGL: 뭔가 허전함...

% >0.5 -> increase diversity from random selection

% Q. how to define a minority instance?
% - the instance with the lowest expected confidence level across all labels.

\subsection{Fine-grained Label-wise Management}
\label{sec:label_management}

Before mixing the two instances by Eq.~\eqref{eq:mixup}, to make noisy multi-labels reliable in support of robust optimization, we perform \emph{label-wise} refinement.

\paragraph{Clean Labels.} To relieve the imbalance problem in label selection, we separately identify clean labels for each class. % and then combine them for the entire data. 
Let $L_{(\tilde{y}_k=1)}$ and $L_{(\tilde{y}_k=0)}$ be the sets of the BCE losses of the positive and negative labels of the $k$-th class,
\begin{equation}
\small
L_{(\tilde{y}_k=l)} = \{{\rm BCE}(f_{(\mathbf{x}, \tilde{y}_k=l)})~ |~ (\mathbf{x}, \tilde{\mathbf{y}}) \in \tilde{\mathcal{D}} ~\wedge~ \tilde{y}_k \in \tilde{\mathbf{y}} ~\wedge~ \tilde{y}_k=l \},\!
\label{eq:partition}
\end{equation}
where $l$ is 1 or 0 for the positive or negative label. 

% =\cup_{(\mathbf{x}, \tilde{\mathbf{y}})\in \tilde{\mathcal{D}}} \{f_{(\mathbf{x}, \tilde{y}_k=1)}\}$
%Let $\tilde{\mathcal{Y}}=[\tilde{\mathbf{y}}_1, \dots, \tilde{\mathbf{y}}_N]^{\top} (\in \{0, 1\}^{N \times K})$ be the noisy label matrix for data of size $N$. We first identify clean labels from the noisy label matrix.
Clean labels exhibit loss values smaller than noise ones due to the memorization effect of DNNs\,\cite{li2020dividemix}. Hence, we fit a bi-modal univariate Gaussian mixture model\,(GMM) to each set of the BCE losses in using the expectation-maximization\,(EM) algorithm, returning $2 \times K$ GMM models for positive and negative labels of $K$ classes, 
\begin{equation}
\small
p_{\mathcal{G}}=\{(\mathcal{G}_{(\tilde{y}_k=1)}, \mathcal{G}_{(\tilde{y}_k=0)})\}_{k=1}^{K}.
\end{equation}
Given the BCE loss of $\mathbf{x}$ for the $k$-th positive or negative label, its clean-label probability is obtained by the posterior probability of the corresponding GMM,
\begin{equation}
\small
\!p_{\mathcal{G}}(\mathbf{x}, \tilde{y}_k=l) = \frac{\mathcal{G}_{(\tilde{y}_k=l)}({\rm BCE}(f_{(\mathbf{x}, \tilde{y}_k=l)})|g) \cdot \mathcal{G}_{(\tilde{y}_k=l)}(g)}{\mathcal{G}_{(\tilde{y}_k=l)}({\rm BCE}(f_{(\mathbf{x}, \tilde{y}_k=l)}))},\!\!\!
\label{eq:clean_p}
\end{equation}
where $g$ denotes a modality for the small-loss\,(clean) label. Thus, a label with $p_{\mathcal{G}}>0.5$ is marked as being clean.

The time complexity of GMM modeling is $\mathcal{O}(NGD)\!=\!\mathcal{O}(N)$ and thus linear to the number of instances $N$, where the number of modalities $G\!=\!2$ and the number of dimensions $D\!=\!1$ (see \cite{trivedi2017handbook} for the proof of the time complexity). Since we model the GMMs once per epoch, the cost involved is expected to be small compared with the training steps of a complex DNN. % JGL: workaround, just added "expected to be", later, if some data is collected, it would be better to add some range (xx--xx\%) 

\paragraph{Re-labeled Labels.} Before the overfitting to noisy labels, a model's prediction delivers useful underlying information on correct labels\,\cite{song2019selfie}. Therefore, we modify the given label if it is not selected as a clean one but the model exhibits high confidence in predictions. To obtain a stable confidence from the model, we ensemble the prediction confidences on two augmented views created by RandAug\,\cite{cubuk2020randaugment}. Given two differently-augmented instances from the original instance $\mathbf{x}$ whose $p_{\mathcal{G}}(\mathbf{x}, \tilde{y}_k) \leq 0.5$, the $k$-th label is re-labeled by
\begin{equation}
\small
\begin{gathered}
1/2 \cdot (f_{({\rm aug}_1(\mathbf{x}), \tilde{y}_k)} + f_{({\rm aug}_2(\mathbf{x}), \tilde{y}_k)}) > \epsilon ~\Longrightarrow~ \tilde{y}_k = 1,\!\!\!\!\\
1/2 \cdot (f_{({\rm aug}_1(\mathbf{x}), \tilde{y}_k)} + f_{({\rm aug}_2(\mathbf{x}), \tilde{y}_k)}) < 1 - \epsilon ~\Longrightarrow~ \tilde{y}_k = 0,\!\!\!\!
\end{gathered}
\label{eq:relabeling}
\end{equation}
where $\epsilon$ is the confidence threshold for re-labeling.

\paragraph{Ambiguous Labels.} The untouched labels, which are neither clean nor re-labeled, are regarded as being ambiguous. These labels are potentially incorrect, but they could hold meaningful information in learning with careful treatment. To squeeze the meaningful information and reduce the potential risk, for these ambiguous labels,
we execute \emph{importance reweighting} which decays a loss based on the clean-label probability estimated by Eq.~\eqref{eq:clean_p}.

\begin{figure*}[t!]
\begin{center}
\includegraphics[width=17.6cm]{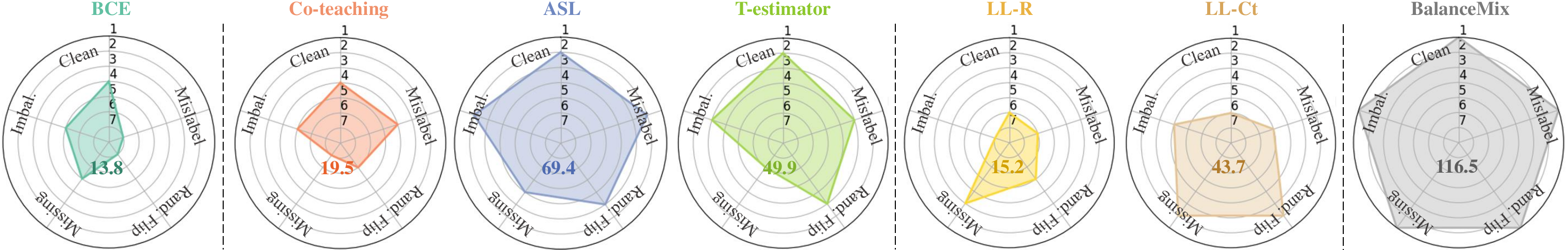}
\end{center}
\vspace*{-0.1cm}
\hspace*{0.4cm} {\small (a) Default.} \hspace*{2.3cm} {\small (b) Methods for Noisy Labels.} \hspace*{2.15cm} {\small (c) Methods for Missing Labels.} \hspace*{1.0cm} {\small (d) Ours.}
\vspace*{-0.15cm}
\caption{Performance ranking (1--7) from five different perspectives. BalanceMix is the \textbf{most versatile} to handle diverse types of label issues in multi-label classification. A number in a pentagon is its area, roughly meaning the overall performance.}
\vspace*{-0.1cm}
\label{fig:ranking}
\end{figure*}
\begin{table*}[t!]
\begin{center}
\footnotesize
\begin{tabular}{L{2.0cm} |L{1.8cm} |X{0.63cm} X{0.63cm} X{0.63cm} |X{0.63cm} X{0.63cm}X{0.63cm} |X{0.63cm} X{0.63cm} X{0.63cm} |X{0.63cm} X{0.63cm} X{0.63cm}}\hline
%\rowcolor{Gray}
\multicolumn{2}{c|}{Class Group} & \multicolumn{3}{c|}{All} & \multicolumn{3}{c|}{Many-shot} & \multicolumn{3}{c|}{Medium-shot} & \multicolumn{3}{c}{Few-shot} \\\toprule
Category & Method & 0\% & 20\% & 40\% & 0\% & 20\% & 40\% & 0\% & 20\% & 40\% & 0\% & 20\% & 40\%  \\\midrule
{Default} 
& BCE Loss & 83.4 & 73.1 & 63.8 & 86.9 & 78.4 & 68.4 & 84.5 & 74.0 & 64.3 & 64.3 & 55.5 & 48.0 \\ \midrule
\multirow{3}{*}{Noisy Labels} 
& Co-teaching  & 82.8 & 82.5 & 78.6 & 87.2 & 87.0 & 82.7 & 84.0 & 83.7 & 79.7 & 61.2 & 60.9 & {54.5} \\ 
& ASL  & \underline{85.0} & \underline{82.8} & {80.3} & \textbf{88.4} & \underline{87.4} & \underline{85.4} & \textbf{86.3} & \underline{84.0} & \underline{81.8} & \underline{67.5} & 62.5 & 55.6 \\
& T-estimator & {84.3} & {82.2} & \underline{80.5} & {87.5} & {86.3} & {85.3} & {85.4} & {83.6} & {81.3} & {67.0} & 59.7 & \underline{61.0} \\ \midrule
\multirow{2}{*}{Missing Labels\!\!} 
& LL-R  & 82.5 & 80.7 & 75.3 & 81.0 & 83.9 & 79.8 & 83.8 & 81.8 & 76.7 & 65.7 & 63.1 & 52.6 \\
& LL-Ct  & 79.4 & 81.3 & 77.4 & 72.3 & 78.8 & 79.3 & 80.5 & 82.6 & 78.8 & 67.0 & \underline{64.3} & {55.9} \\ \midrule
Proposed & BalanceMix & \textbf{85.2} & \textbf{84.3} & \textbf{81.6} & \textbf{88.4} & \textbf{87.7} & \textbf{85.5} & \underline{86.1} & \textbf{85.2} & \textbf{82.6} & \textbf{70.2} & \textbf{68.7} & \textbf{63.1} \\ \bottomrule
\end{tabular}
\end{center}
\vspace*{-0.4cm}
\caption{Last mAPs on MS-COCO with \textbf{mislabeling} of $0$--$40\%$. The 1st and 2nd best values are in bold and underlined.} 
\label{table:coco_mislabeling}
\vspace*{-0.45cm}
\end{table*}

\vspace*{-0.05cm}
\subsection{Optimization with BalanceMix}
Given two instances sampled from the random and minority samplers, their multi-labels are first refined by the label-wise management setup. The \emph{reliability} of each refined label is stored as {\sf C} for clean labels, {\sf R} for re-labeled labels, and {\sf U} for ambiguous labels.
Then, a minority-augmented instance is generated by Mixup with the mixed multi-labels. 
The reliability of each label of the augmented instance follows that of the instance selected by the random sampler, because it always dominates in mixing by $\lambda\geq0.5$ in Eq.\ \eqref{eq:mixup}. The loss function of \algname{} is defined on the minority-augmented mini-batch $B_{mix}$ by 
\begin{equation}
\small
\begin{gathered}
\mathcal{L}_{\rm ours}(B_{mix}; \Theta) = \frac{1}{|B_{mix}|} \!\!\!\!\!\!\!\!\!\sum_{~~~~~~~(\mathbf{x}^{mix}, \tilde{\mathbf{y}}^{mix})\in B_{mix}} \!\!\!\!\!\!\!\!\!\!\!\!\!\!\!\!\ell(\mathbf{x}^{mix}, \tilde{\mathbf{y}}^{mix}), \\
%\ell(\mathbf{x}, \mathbf{y}) = \!\!\!\sum_{k \in {\sf C} \cup {\sf R}} \!\!\!{\rm BCE}(f_{(\mathbf{x}, \tilde{y}_k)}) + \!\sum_{k \in {\sf U}} p_{\mathcal{G}}(\mathbf{x}, \tilde{y}_k)\!\cdot\! {\rm BCE}(f_{(\mathbf{x}, \tilde{y}_k)}). \!\!\!\!
{\rm where} ~~\ell(\mathbf{x}^{mix}, \mathbf{y}^{mix}) = \!\!\!\sum_{k \in {\sf C} \cup {\sf R}} \!\!\!{\rm BCE}(f_{(\mathbf{x}^{mix}, \tilde{y}_k^{mix})}) \\+ \sum_{k \in {\sf U}} p_{\mathcal{G}}(\mathbf{x}^{mix}, \tilde{y}_k^{mix})\!\cdot\! {\rm BCE}(f_{(\mathbf{x}^{mix}, \tilde{y}_k^{mix})}). \!\!\!\!
\end{gathered}
\label{eq:update_our}
\end{equation}
We perform standard training for warm-up epochs and then apply the proposed loss function in Eq.\ \eqref{eq:update_our}. 

% Clean Label Set of "C", 
% Re-labeled Label Set of "R" => two augmented view -> ensemble prediction confidence > threshold.
% Ambiguous Label Set of "U" => the rest unclean labels

% Final loss = \Sum { BCE() if label in "C" \cup "R", clean_p(l) * BCE() if label in "U" }

%% file: 4-experiment.tex
\section{Evaluation}
\label{sec:evaluation}

\paragraph{Datasets.}
% dataset and
% We conduct multi-label classification on three benchmark datasets. 
Pascal-VOC\,\cite{everingham2010pascal} and MS-COCO\,\cite{lin2014microsoft} are the most widely-used datasets with well-curated labels of 20 and 80 common classes. In contrast, DeepFashion\,\cite{liuLQWTcvpr16DeepFashion} is a real-world in-shopping dataset with noisy weakly-annotated labels for $1,000$ descriptive attributes. %We use a fine-grained subset of DeepFashion with 16,000 training and 4,000 validation instances as well as multi-labels of $26$ attribute classes, which are provided by the authors. 

\begin{table*}[t!]
\begin{center}
\footnotesize
\begin{tabular}{L{2.0cm} |L{1.8cm} |X{0.63cm} X{0.63cm} X{0.63cm} |X{0.63cm} X{0.63cm} X{0.63cm} |X{0.63cm} X{0.63cm} X{0.63cm} |X{0.63cm} X{0.63cm} X{0.63cm}}\hline
%\rowcolor{Gray}
\multicolumn{2}{c|}{Class Group} & \multicolumn{3}{c|}{All} & \multicolumn{3}{c|}{Many-shot} & \multicolumn{3}{c|}{Medium-shot} & \multicolumn{3}{c}{Few-shot} \\\toprule
Category & Method & 0\% & 20\% & 40\% & 0\% & 20\% & 40\% & 0\% & 20\% & 40\% & 0\% & 20\% & 40\%  \\\midrule
{Default} 
& BCE Loss & 83.4 & 59.8 & 43.5 & 86.9 & 71.2 & 65.1 & 84.5 & 60.6 & 43.2 & 64.3 & 38.9 & 30.6 \\ \midrule
\multirow{3}{*}{Noisy Labels} 
& Co-teaching  & 82.8 & 65.5 & 43.6 & 87.2 & 76.1 & 61.9 & 84.0 & 66.8 & 43.8 & 61.2 & 38.3 & 25.9 \\ 
& ASL  & \underline{85.0} & \underline{75.0} & 66.2 & \textbf{88.4} & \underline{84.4} & \textbf{82.8} & \textbf{86.3} & \underline{77.0} & 67.7 & \underline{67.5} & 39.2 & 30.8 \\
& T-estimator  & {84.3} & {74.3} & \underline{69.9} & {87.5} & {82.6} & {80.8} & {85.4} & {76.0} & \underline{71.5} & {67.4} & 43.6 & 39.1 \\\midrule
\multirow{2}{*}{Missing Labels\!\!} 
& LL-R  & 82.5 & 74.0 & {69.3} & 81.0 & 77.2 & 76.6 & 83.8 & 75.8 & 71.0 & 65.7 & \underline{46.0} & 38.6 \\
& LL-Ct  & 79.4 & 73.2 & {70.1} & 72.3 & 75.5 & 76.5 & 80.5 & 75.0 & {71.8} & {67.0} & {45.5} & \underline{41.1} \\ \midrule
Proposed & BalanceMix & \textbf{85.2} & \textbf{76.5} & \textbf{74.5} & \textbf{88.4} & \textbf{84.5} & \underline{81.3} & \underline{86.1} & \textbf{78.2} & \textbf{76.3 }& \textbf{70.2} & \textbf{46.1} & \textbf{43.0} \\ \bottomrule
\end{tabular}
\end{center}
\vspace*{-0.4cm}
\caption{Last mAPs on MS-COCO with \textbf{random flipping} of $0$--$40\%$. The 1st and 2nd best values are in bold and underlined.} 
\label{table:coco_random_flip}
\vspace*{-0.2cm}
\end{table*}

\begin{table*}[t]
\begin{center}
\footnotesize
\begin{tabular}{L{2.0cm} |L{2.2cm} |X{1.0cm} X{1.0cm} X{1.0cm} X{1.0cm} |X{1.0cm} X{1.0cm} X{1.0cm}}\hline
\multicolumn{2}{c|}{Datasets} & \multicolumn{4}{c|}{MS-COCO} & \multicolumn{3}{c}{Pascal-VOC} \\\toprule
Category & Method & All & Many & \!\!\!Medium\!\!\! & Few & All & \!\!\!Medium\!\!\! & Few \\\midrule
%Full Labels & BalanceMix & 85.2 & 88.4 & 86.1 & 70.2 & 93.3 & 95.1 & 92.5 \\ \hline\midrule
{Default} 
& BCE Loss & 69.7 & 71.7 & 70.6 & 54.4 & 85.7 & 89.2 & 84.2\\ \midrule
\multirow{3}{*}{Noisy Labels} 
& Co-teaching  & 68.1 & 61.5 & 69.2 & {59.1} & 80.9 & 87.2 & 78.1  \\ 
& ASL  & 73.3 & \textbf{77.7} &  74.7 & 49.2 & 86.8 & 82.1 & {88.8} \\
& T-estimator & {16.8} & {43.2} & {16.3} & {3.3} & 86.2 & 88.9 & 85.0 \\\midrule
\multirow{2}{*}{Missing Labels} 
& LL-R  & 74.2 & 75.4 & 75.3 & 58.7 & {89.1} & \underline{91.5} & 88.1\\
& LL-Ct   & \underline{76.9} &  \underline{77.4} &  \underline{78.2} & 57.6 &  \underline{89.3} & \underline{91.5} & 88.3\\ \midrule
Proposed & BalanceMix & \textbf{77.4} & 76.2 & \textbf{78.5} & \textbf{61.3} & \textbf{92.6} & \textbf{94.5} & \textbf{91.8}\\ \bottomrule
\end{tabular}
\end{center}
\vspace*{-0.4cm}
\caption{Last mAPs on MS-COCO and Pascal-VOC in the \textbf{missing label\,(single positive label)} setup.} 
\vspace*{-0.5cm}
\label{table:coco_single_label}
\end{table*}

\paragraph{Imbalanced and Noisy Labels.} 
The three datasets contain different levels of natural imbalance. Pascal-VOC, MS-COCO, and DeepFashion have the class imbalance ratios\footnote{The ratio of the number of the instances in the most frequent class to that of the instances in the least frequent class.} of $14$, $339$, and $239$, and the positive-negative imbalance ratios of $13$, $27$, and $3$, respectively. See the detailed analysis of the imbalance in Appendix C.
We artificially contaminate Pascal-VOC and MS-COCO to add three types of label noise. First, for mislabeling, we inject class-dependent label noise. Given a noise rate $\tau$, the presence of the $i$-th class is mislabeled as that of the $j$-th class with a probability of $\rho_{i \rightarrow j}$; we follow the protocol used for a long-tail noisy label setup\,\cite{wei2021robust}. For the two different classes $i$ and $j$, 
\begin{equation}
\small
\rho_{i \rightarrow j} = p(\tilde{y}_{i}=0, \tilde{y}_{j}\!=\!1 | y_{i}\!=\!1) = \tau \cdot N_j / (N - N_i),
\end{equation}
where $N_i$ is the number of positive labels for the $i$-th class. % This prevents class changes under class imbalance. 
Second, for random flipping, all positive and negative labels are flipped independently with the probability of $\tau$. Third, for missing labels, we follow the single positive label setup\,\cite{kim2022large}, where one positive label is selected at random and the other positive labels are dropped.% (i.e., treated as negative labels).% For the two former label noise, we test three noise rates $\tau \in \{0.0, 0.2, 0.4\}$.

\paragraph{Algorithms.}
We use the ResNet-50 backbone pre-trained on ImageNet-1K and fine-tune using SGD with a momentum of 0.9 and resolution of $448\times448$. We compare \algname{} with a standard method using the BCE loss\,(Default) and \emph{five} state-of-the-art methods, categorized into two groups.
The former is to handle \emph{noisy} labels based on instance-level selection, loss reweighting, and noise transition matrix estimation---Co-teaching\,\cite{han2018co}, ASL\,\cite{ben2020asymmetric}, and T-estimator\,\cite{li2022estimating}. 
% \footnote{For T-estimator, as in the original paper, we use $10$\% of noisy training data as the noisy validation set for estimating a noise transition matrix.}
The latter is to handle \emph{missing} labels based on label rejection and correction---LL-R and LL-Ct\,\cite{kim2022large}. For data augmentation, we apply RandAug and Mixup to all methods, except Default using only RandAug. The results of Default with Mixup are presented in Table \ref{table:com_analysis}.  % {\color{purple}We contrast the performance gain of using Mixup + random sampler and Mixup + minority sampler on top of the Default method in Table \ref{table:com_analysis}.}

As for our hyperparameters, the coefficient $\alpha$ for Mixup is set to be $4.0$; and the confidence threshold $\epsilon$ for re-labeling is set to be $0.975$ for the standard, mislabeling, and random flipping settings with \emph{multiple} positive labels, but it is set to be $0.550$ for the missing label setting with a \emph{single} positive label. More details of configuration and hyperparameter search can be found in Appendices D and E. 
% The source code is available at {\url{bit.ly/3DM4Qet}}. % r

\paragraph{Evaluation Metric.} 
We report the overall validation\,(or test) mAP at the last epoch over three disjoint class subsets: many-shot (more than 10,000 positive labels), medium-shot (from 1,000 to 10,000 positive labels), and few-shot (less than 1,000 positive labels) classes. The result at the last epoch is commonly used in the literature on robustness to label noise\,\cite{han2018co}. We repeat every task thrice, and see Appendix F for the standard error.

\vspace*{-0.1cm}
\subsection{Overall Analysis on Five Perspectives}

Fig.~\ref{fig:ranking} shows the overall performance rankings aggregated\footnote{For each perspective, we respectively compute the ranking on each dataset and then sum up the rankings to get the final one.} on Pascal-VOC and MS-COCO for five different perspectives: ``Clean'' for when label noise is not injected, ``Mislabel'' for when labels are mislabeled with the noise ratio of $20$--$40\%$, ``Rand.~Flip'' for when labels are randomly flipped with the noise ratio of $20$--$40\%$, ``Missing'' for when the single positive label setup is used, and ``Imbal.'' for when few-shot classes without label noise are used.

Only \algname{} operates in all scenarios with high performance: its minority-augmented mixing overcomes the problem of \emph{imbalanced labels} while its fine-grained label-wise management adds robustness to \emph{diverse types of label noise}. Except \algname{}, the five existing methods have pros and cons. The three methods of handling noisy labels in Fig.~\ref{fig:ranking}(b) generally perform better for mislabeling and random flipping than the others; but the instance-level selection of Co-teaching is not robust to random flipping where a significant number of negative labels are flipped to positive ones. In contrast, the two methods of handling missing labels in Fig.~\ref{fig:ranking}(c) perform better with the existence of missing labels than Co-teaching, ASL, and T-estimator. LL-Ct (label correction) is more suitable than LL-R (label rejection) for mislabeling and random flipping since label correction has a potential to re-label some of incorrect labels. For the imbalance, ASL shows reasonable performance on the few-shot subset by adopting a modified focal loss.

\subsection{Results on Imbalanced and Noisy Labels}

We evaluate the performance of \algname{} on MS-COCO with three types of synthetic label noise and on DeepFashion with \emph{real-world} label noise. We defer the results on Pascal-VOC to Appendix F for the sake of space.

\subsubsection{Mislabeling (Table \ref{table:coco_mislabeling}).} % summarizes the last mAPs on MS-COCO with mislabeling noise. % We add two components of minority-augmented mixing and fine-grained label management on top of Default\,(BCE). 
% The improvements of them are promising; 
% The improvements of \algname{} over the baseline BCE are remarkable. 
\algname{} achieves not only the best overall mAP (see the ``All'' column) with varying mislabeling ratios, but also the best mAP on few-shot classes (see the ``Few-shot'' column). It shows higher robustness even compared with the three methods designed for noisy labels. ASL performs well among the compared methods, but its weighting scheme of pushing higher weights to difficult-to-learn labels could lead to overfitting to difficult incorrect labels; hence, when the noise ratio increases, its performance rapidly degrades from $67.5\%$ to $55.6\%$ in the few-shot classes. Both methods for missing labels perform better than the default method\,(BCE), but are still vulnerable to mislabeling.

\begin{table}[t]
\vspace*{-0.0cm}
\begin{center}
\footnotesize
\begin{tabular}{L{2.0cm} |X{1.05cm} |X{1.05cm} X{1.05cm} X{1.05cm}}\hline
Class Group & All & Many & \!\!Medium\!\! & Few \\\toprule
BCE Loss & 75.2 & 93.4 & 84.4 & 53.4\\ 
Co-teaching  & 66.8 & 90.7 & 81.3 & 32.8 \\
ASL  & {76.4} & {94.4} & {85.2} & {55.4} \\
T-estimator  & {75.4} & {94.7} & {84.8} & 53.1 \\
LL-R & 75.3 & 93.3 & 84.2 & 53.8 \\
LL-Ct  & 75.2 & 92.6 & 84.2 & 53.8 \\
BalanceMix & \textbf{77.0} & \textbf{95.2} & \textbf{85.6} & \textbf{56.4} \\ \bottomrule
\end{tabular}
\end{center}
\vspace*{-0.4cm}
\caption{mAPs on DeepFashion with \textbf{real-world} noisy multi-labels using seven multi-label classification methods.} 
\label{table:deepfashion_real}
\vspace*{-0.55cm}
\end{table}

\begin{table*}[t!]
\begin{center}
\footnotesize
\begin{tabular}{L{4.5cm} |X{1.9cm} |X{1.9cm} |X{1.9cm} |X{1.9cm} ||X{2.2cm}  }\hline
Component\!\!           &  Clean Label & Mislabel 40\% & \!\!Rand Flip 40\%\!\! & Missing Label & Overall (Mean) \\\toprule
Default (BCE Loss)                   & \!\!\!\!\!\!\!\!\!\!\!\!\!\!\!\!\!\!83.4 & \!\!\!\!\!\!\!\!\!\!\!\!\!\!\!\!\!\!63.3 & \!\!\!\!\!\!\!\!\!\!\!\!\!\!\!\!\!\!43.0 & \!\!\!\!\!\!\!\!\!\!\!\!\!\!\!\!\!\!72.6 & \!\!\!\!\!\!\!\!\!\!\!\!\!\!\!\!\!\!65.6 \\ 
$+$ Random Sampler ($\approx$ Mixup)  & 84.2 ($+$0.8) & 67.4 ($+$3.3) & \,\,\,64.9 ($+$21.9) & 73.3 ($+$0.7) & 72.5 ($+$6.9) \\ \midrule
$+$ Minority Sampler (in Eq.\,\eqref{eq:weights})    & 85.1 ($+$1.7) & 70.2 ($+$6.9) & \,\,\,67.2 ($+$24.2) & 74.2 ($+$1.6) & 74.2 ($+$8.6) \\
$+$ Clean Labels (in Eq.\,\eqref{eq:clean_p})        & 84.9 ($-$0.2) & 76.1 ($+$5.9) & \textbf{74.9} ($+$7.7) & 74.6 ($+$0.4) & 77.6 ($+$3.4) \\
$+$ Re-labeled Labels (in Eq.\,\eqref{eq:relabeling})   & \textbf{85.3} ($+$0.4) & 80.2 ($+$3.9) & \textbf{74.9} ($+$0.0) & 76.1 ($+$1.5) & 79.1 ($+$1.5)\\ 
$+$ Ambiguous Labels  (in Eq.\,\eqref{eq:update_our})    & \textbf{85.3} ($+$0.0) & \textbf{81.6} ($+$1.4) & 74.5 ($-$0.4) & \textbf{77.4} ($+$1.3) & \textbf{79.7} ($+$0.6) \\ \bottomrule
\end{tabular}
\end{center}
\vspace*{-0.4cm}
\caption{Component analysis of \algname{} on MS-COCO. The values in parentheses are the gain caused by each component.} 
\label{table:com_analysis}
\vspace*{-0.45cm}
\end{table*}

\subsubsection{Random Flipping (Table \ref{table:coco_random_flip}).} %summarizes the last mAPs on MS-COCO with random flipping noise, 
This is more challenging than mislabeling noise, in considering that even negative labels are flipped by a given noise ratio. Accordingly, the mAP of Co-teaching and ASL drops significantly when the noise ratio reaches $40\%$ (see the ``All'' column), which implies that instance selection in Co-teaching and loss reweighting in ASL are ineffective to overcome random flipping. T-estimator shows a better result at the noise ratio of $40\%$ than ASL by estimating the noise transition matrix per class. Overall, \algname{} achieves higher robustness against a high flipping ratio of $40\%$ with fine-grained label-wise management; its performance drops by only $10.7\%p$, which is much smaller than $39.2\%p$, $18.8\%p$, and  $14.4\%p$ of Co-teaching, ASL, and T-estimator, respectively. Thus, it maintains the best mAP for all class subsets in general.
%its performance drops by only $10.7\%p$ (from $85.2\%$ to $74.5\%$), which is much smaller than $39.2\%p$ (from $82.8\%$ to $43.6\%$), $18.8\%p$ (from $85.0\%$ to $66.2\%$), and  $14.4\%p$ (from $84.3\%$ to $69.9\%$) of Co-teaching, ASL, and T-estimator, respectively. Therefore, it maintains the best mAP for all class subsets in general.

\subsubsection{Missing Labels (Table \ref{table:coco_single_label}).} %summarizes the last mAPs on MS-COCO and Pascal-VOC with missing labels. 
Unlike the mislabeling and random flipping, LL-R and LL-Ct generally show higher mAPs than the methods for noisy labels, because LL-R and LL-Ct are designed to reject or re-label unobserved positive labels that are erroneously considered as negative ones. Likewise, the label-wise management of \algname{} includes the re-labeling process, fixing incorrect positive and negative labels to be correct. In addition, it shows higher mAP in the few-shot classes than LL-Ct due to the consideration of imbalanced labels. Thus, it consistently maintains its performance dominance. 
{Meanwhile, T-estimator performs badly in MS-COCO due to the complexity of transition matrix estimation.}
%, where all except one positive label are missing, but works fairly well in Pascal-VOC since there are only 1 or 2 labels originally.}

\subsubsection{Real-world Noisy Labels (Table \ref{table:deepfashion_real}).} A real-world noisy dataset, DeepFashion, likely contains \emph{all} the label noises---mislabeling, random flipping, and missing labels---along with class imbalance. % In Table \ref{table:deepfashion_real}, \algname{} shows the best mAP in all class subsets, while ASL shows the second best mAP.
%Interestingly, the order of the methods in terms of the mAP in Table \ref{table:deepfashion_real} is similar to that of the methods in terms of the size of a pentagon in Fig. \ref{fig:ranking}. Since Fig. \ref{fig:ranking} is obtained by {aggregating} the results from each {individual} noise type, we confirm that a real-world noisy dataset (e.g., DeepFashion) encompasses these diverse types together. 
Therefore, our motivation for a holistic approach is of importance for real use cases.

The relatively small performance gain is attributed to a small percentage (around 8\%) of noise labels\,\cite{song2022learning} in DeepFashion, because its fine-grained labels were annotated via a crowd-sourcing platform which can be relatively reliable. The performance gain will increase for datasets with a higher noise ratio. %owing to unreliable annotation. 
% The relatively small performance gain can be attributed to the fact that DeepFashion's fine-grained labels were annotated via a crowd-sourcing platform, which resulted in a minor percentage (around 8\%) of noise labels \cite{song2022learning}. That is, it includes realistic noisy labels, but its noisy ratio is much lighter than the synthetic case of 40\%.

% Area: BalanceMix ASL LL-Ct Co-teaching LL-R BCE
% mAT: BalanceMix ASL T-estimator LL-CT Co-teaching LL-R BCE
\vspace*{-0.1cm}
\subsection{Component Ablation Study}

We conduct a component ablation study by adding the main components one by one on top of the default method.  Table \ref{table:com_analysis} summarizes the mAP and average performance of each of five variants. The first variant of using only a random sampler is equivalent to the original Mixup. 

First, using only a random sampler like Mixup does {\rm not} sufficiently improve the model performance, but adding the minority sampler achieves sufficient improvement because it takes imbalanced labels into account.
% the use of Mixup with a random sampler improves model performance and robustness, but provides a greater improvement when replacing the random sampler with our minority sampler because it takes into account imbalanced labels,} i.e., the huge mAP gain for few-shot classes in all scenarios on MS-COCO (see the ``Few-shot'' column in Tables \ref{table:coco_mislabeling}--\ref{table:coco_single_label} for the three different noisy label setups). 
%
Second, exploiting only the selected clean labels increases the mAP when positive labels are corrupted with mislabeling and random flipping. However, this approach is not that beneficial in the clean and missing label setups, where all positive labels are regarded as being clean; it also simply discards all (expectedly) unclean negative labels without any treatment.
%becuase it discards all the (expected) unclean labels. 
Third, re-labeling complements the limitation of clean label selection, providing additional mAP gains in most scenarios. Fourth, using ambiguous labels adds further mAP improvement except for the random flipping setup.

In summary, since all the components in \algname{} generally add a synergistic effect, leveraging all of them is recommended for use in practice. {In Appendix G, we (1) analyze the impact of minority-augmented mixing on diversity changes, (2) provide the pure effect of label-wise management, and (3) report its accuracy in selecting clean labels and re-labeling incorrect labels.}

\begin{table}[t]
\begin{center}
\footnotesize
\begin{tabular}{L{2.1cm} |X{1.7cm} |X{1.5cm} |X{1.4cm}}\hline
Method & Backbone & Resolution & mAP (All) \\\toprule
%ML-GCN & ResNet-101 & 448$\times$448 & 83.0 \\ 
%KSSNET & ResNet-101 & 448$\times$448 & 83.7 \\ 
MS-CMA  & ResNet-101 & 448$\times$448 & 83.8 \\ 
ASL     & ResNet-101 & 448$\times$448 & 85.0 \\ 
ML-Decoder & ResNet-101 & 448$\times$448 & 87.1 \\
\textbf{BalanceMix} & ResNet-101 & 448$\times$448 & \textbf{87.4\,(+0.3)} \\ \midrule
ASL \   & TResNet-L & 448$\times$448 & 88.4 \\ 
Q2L     & TResNet-L & 448$\times$448 & 89.2 \\ 
ML-Decoder & TResNet-L & 448$\times$448 & 90.0 \\
\textbf{BalanceMix} & TResNet-L & 448$\times$448 & \textbf{90.5\,(+0.5)} \\ \midrule
ML-Decoder  & TResNet-L & 640$\times$640 & 91.1 \\
ML-Decoder  & TResNet-XL & 640$\times$640 & 91.4 \\
\textbf{BalanceMix}\!\! & TResNet-L & 640$\times$640 & \textbf{91.7\,(+0.6)} \\\bottomrule
\end{tabular}
\end{center}
\vspace*{-0.4cm}
\caption{\textbf{State-of-the-art comparison} on MS-COCO. The values in parentheses are the improvements over the latest method using the same backbone.} 
\label{table:sota_coco}
\vspace*{-0.50cm}
\end{table}

\vspace*{-0.1cm}
\subsection{State-of-the-art Comparison on MS-COCO}

We compare \algname{} with several methods showing the state-of-the-art performance with a ResNet backbone on MS-COCO. The results are borrowed from Ridnik et al.\,\cite{ridnik2023ml}, and we follow exactly the same setting in backbones, image resolution, and data augmentation. BalanceMix is implemented on top of ML-Decoder for comparison. All backbones are pre-trained on ImageNet. The compared methods are developed without consideration of label noise, but we find out that MS-COCO originally has noisy labels (see Appendix H for examples). 

Table \ref{table:sota_coco} summarizes the best mAP on MS-COCO without synthetic noise injection. For the $448\times448$ resolution, \algname{} improves the mAP by $0.3$--$0.5\%p$ when using ResNet-101 and TResNet-L. For the $640\times640$ resolution, its improvement over ML-Decoder becomes $0.6\%p$ when using TResNet-L. The $91.7$mAP of \algname{} with TResNet-L is even higher than the 91.4mAP of ML-Decoder with TResNet-XL. %This result is a \emph{record-breaking} mAP for multi-label classification on MS-COCO according to Papers with Code\footnotemark[3] at the time of submission. 
%This new record on MS-COCO indeed assures the versatility of \algname{}. % for real imbalanced and noisy labels. % seems to be verbose
% \footnote{Although ADDS\,\cite{xu2022adds} using ViT-L-336 set the record in August 2022, it is not officially published and is hard to be reproduced because its source code is not released at the time of submission. In addition, ViT-L has 307M learnable parameters, which is much larger than 54.7M of TResNet-L.}

%% file: 5-conclusion.tex
\vspace*{-0.1cm}
\section{Conclusion}
\label{sec:conclusion}
\vspace*{-0.1cm}

We propose {BalanceMix}, which can handle imbalanced labels and diverse types of label noise. The minority-augmented mixing allows for adding sparse context in minority classes to majority classes without losing diversity. The label-wise management realizes a robust way of exploiting noisy multi-labels without overfitting. %, categorizing them into clean, re-labeled, and ambiguous labels. 
Through experiments using real-world and synthetic noisy datasets, we verify that \algname{} outperforms state-of-the-art methods in each setting of mislabeling, flipping, and missing labels, with the co-existence of severe class imbalance.  
%In particular, it sets a new record of 91.7mAP on MS-COCO. 
Overall, this work will inspire subsequent studies to handle imbalanced and noisy labels in a holistic manner.

\section*{Acknowledgements}
This work was supported by Institute of Information \& Communications Technology Planning \& Evaluation\,(IITP) grant funded by the Korea government\,(MSIT) (No. 2020-0-00862, DB4DL: High-Usability and Performance In-Memory Distributed DBMS for Deep Learning). Additionally, this work was partly supported by the FOUR Brain Korea 21 Program through the National Research Foundation of Korea (NRF-5199990113928).

%% file: 6-appendix.tex
%\begin{figure*}[t!]
%\begin{center}
%\includegraphics[width=17.5cm]{figures/appendix/sampling.pdf}
%\end{center}
%\vspace*{-0.15cm}
%\hspace*{2.55cm}{\small (a) Clean Labels} \hspace*{3.4cm}{\small (b) Mislabeling 40\%} %\hspace*{2.9cm}{\small (c) Single Positive Label}
%\vspace*{-0.15cm}
%\caption{Maximum and minimum sampling probabilities of the minority sampler over training epochs on MS-COCO.}
%\vspace*{-0.1cm}
%\label{fig:sampling_p}
%\end{figure*}

\section{A. Pseudocode}

The overall procedure of \algname{} is described in Algorithm \ref{alg:proposed_algorithm}, which is simple and self-explanatory. During the warm-up phase, it updates the model with a standard approach using the BCE loss on the minority-augmented mini-batch. After the warm-up phase, fine-grained label-wise management is performed before generating the minority-augmented mini-batch; in detail, all labels are processed and categorized into clean, re-labeled, and ambiguous ones. Next, the two mini-batches are mixed by Eq.~(5) with the refined labels. Then, the model is updated by the proposed loss function in Eq.~(10).

\setlength{\textfloatsep}{5pt}% Remove \textfloatsep
\begin{algorithm}[h!]
\caption{BalanceMix}
\label{alg:proposed_algorithm}
\begin{algorithmic}[1]
\REQUIRE {$\tilde{\mathcal{D}}$: noisy data, $b$: batch size, $epoch$: training epochs $warm$: warm-up epochs, $\alpha$: mixup coefficient, $\epsilon$: re-labeling threshold}
\ENSURE {$\Theta_t$: DNN parameters}
\INDSTATE[0] {$\Theta \leftarrow \text{Initialize DNN parameters}$};
\INDSTATE[0] {\bf for} $i=1$ {\bf to} $epoch$ {\bf do} 
\INDSTATE[1] {\bf for} $j=1$ {\bf to} $|{\tilde{\mathcal{D}}}|/b$ {\bf do}
\INDSTATE[2] {\COMMENT{{Sampling from two samplers}}}
\INDSTATE[2] {Draw a mini-batch ${B}_{R}$ by the random sampler;}
\INDSTATE[2] {Draw a mini-batch ${B}_{M}$ by the minority sampler;}
\INDSTATE[2] {\bf if} $i \leq warm$ {\bf then}
\INDSTATE[3] {\COMMENT{{ Update with given labels}}}
\INDSTATE[3] {Generate a mini-batch ${B}_{mix}$ by Eq.~(5);}
\INDSTATE[3] {Update the model by Eq.~(1);}
\INDSTATE[2] {\bf else}
\INDSTATE[3] {\COMMENT{{Update with label management}}}
\INDSTATE[3] Perform the label-wise management;
\INDSTATE[3] Generate a mini-batch ${B}_{mix}$ by Eq.~(5);
\INDSTATE[3] Update the model by Eq.~(10);
\INDSTATE[1] {\COMMENT{{Update the minority sampler and GMMs}}}
\INDSTATE[1] Update the sampling probability by Eq.~(4);
\INDSTATE[1] Fitting the GMMs to the loss of entire data;
\INDSTATE[0] {\bf return} $\Theta$
\end{algorithmic}
\end{algorithm} 
\vspace*{-0.3cm}

%\subsubsection{Complexity of GMM.} We optimize GMM models via the EM algorithm. 
%https://lear.inrialpes.fr/~verbeek/papers/verbeek03neco.pdf
% the time complexity of GMM modeling is $\mathcal{O}(T(NGD^{2}+GD^{3}))=\mathcal{O}(N)$
% https://www.sciencedirect.com/topics/computer-science/gaussian-mixture-model

\section{B. Analysis of Minority Sampling}

\begin{figure}[t!]
\begin{center}
\includegraphics[width=8.3cm]{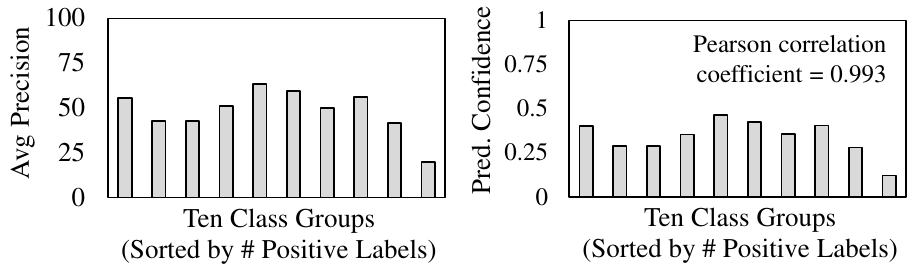}
\end{center}
\vspace*{-0.4cm}
\caption{Prediction confidence (left) and average precision (right) in COCO with {mislabeling} of $40\%$ at the 40$\%$ of training epochs.}
\label{fig:noisy_corr}
\vspace*{-0.2cm}
\end{figure}

We analyze the correlation between average precision and prediction confidence in the presence of noisy labels. Figure 3 was obtained without label noise. Figure \ref{fig:noisy_corr} is obtained \emph{with} label noise. The {Perason correlation coefficient} is still very high, though the absolute values of the confidence and precision are decreased owing to label noise. The coefficient was calculated {between ten class groups}. 

\begin{figure}[t]
\centering
\includegraphics[width=5.3cm]{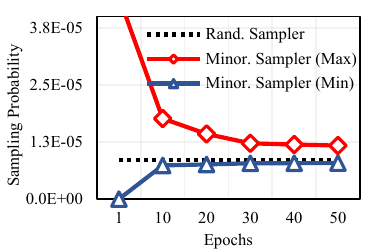}
\vspace*{-0.3cm}
\caption{Sampling probability over the training period.}
\vspace*{0.0cm}
\label{fig:sampling_analysis}
\end{figure}

In addition, we show how the sampling probability changes with our confidence-based minority oversampling method in Figure \ref{fig:sampling_analysis}. Minority instances are initially oversampled with a high probability, but the degree of oversampling gradually decreases as the imbalance problem gets resolved.

\section{C. Imbalance in Benchmark Datasets}

\begin{figure*}[t!]
\centering
\includegraphics[width=17.5cm]{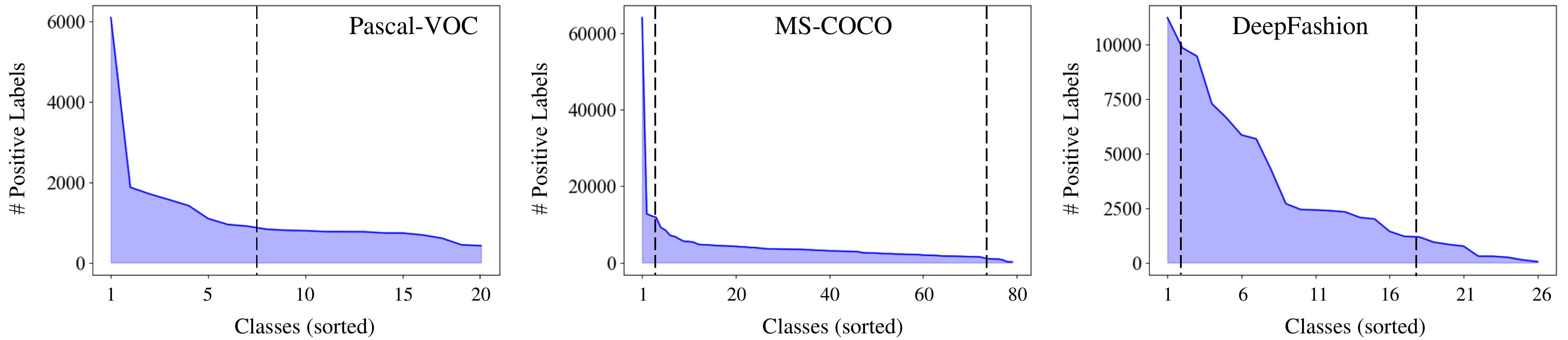}
\vspace*{-0.3cm}
\caption{Imbalanced labels of the three benchmark datasets.}
\vspace*{-0.15cm}
\label{fig:imb_analysis}
\end{figure*}

We investigate the imbalance of positive labels across classes in three benchmark datasets, Pascal-VOC\footnote{\url{http://host.robots.ox.ac.uk/pascal/VOC/}}, MS-COCO\footnote{\url{https://cocodataset.org/}}, and DeepFashion\footnote{\url{https://mmlab.ie.cuhk.edu.hk/projects/DeepFashion.html}}. We use a fine-grained subset of DeepFashion with 16,000 training and 4,000 validation instances as well as multi-labels of $26$ attribute classes, which are provided by the authors. Fig.~\ref{fig:imb_analysis} shows the distribution of the numbers of positive labels across classes, where the dashed lines split the classes into many-shot $[10000,\infty)$, medium-shot $[1000, 10000)$, and  few-shot $[0, 1000)$ classes; Pascal-VOC does not have the many-shot classes. 

A few majority classes occupy most of the positive labels in the data. Hence, we define the \emph{class imbalance ratio} following the literature\,\cite{zhang2021distribution, park2022majority},
\begin{equation}
{\rm CLS\_Imb.} = {\rm max}_{1\leq i\leq k} N_i / {\rm min}_{1\leq i\leq k} N_i,
\end{equation}
which is the maximum ratio of the number of positive labels in the majority class to that in the minority class.
In addition, an image contains few positive labels but many negative labels. Hence, we define the  \emph{positive-negative ratio} by
\begin{equation}
{\rm PN\_Imb.} = \sum_{1\leq i\leq k} N_i^{\prime} / \sum_{1\leq i\leq k} N_i,
\end{equation}
where $N_i^\prime$ is the number of negative labels for the $i$-th class.
As for these two imbalance ratios, Pascal-VOC, MS-COCO, and DeepFashion have class imbalance ratios of $14$, $339$, and $239$, and positive-negative imbalance ratios of $13$, $27$, and $3$, respectively. 

\begin{table*}[t!]
\begin{minipage}{\columnwidth}
\begin{center}
\footnotesize
\begin{tabular}{X{0.8cm} |X{1.5cm} X{1.5cm} ||X{0.8cm} |X{1.5cm}}\hline
$\epsilon$ & \!\!\!\!\!\!\!\!\!Mis. $20\%$\!\!\!\!\!\!\!\!\! & \!\!\!\!\!\!\!\!\!Mis. $40\%$\!\!\!\!\!\!\!\!\! & $\epsilon$ & \!\!\!\!\!\!Missing Label\!\!\!\!\!\! \\\toprule
\!1.000\!\! & 82.5 & 77.8 & \!\!0.550\!\! & \textbf{77.4}\\ 
\!0.975\!\!       & \textbf{84.3} & 81.4 & \!\!0.600\!\! & 77.3 \\
\!0.950\!\!       & 84.0 & \textbf{81.5} & \!\!0.700\!\! & 76.8 \\
\!0.900\!\!       & 83.7 & 81.3 & \!\!0.900\!\! & 76.3 \\ \bottomrule
\end{tabular}
\end{center}
\vspace*{-0.25cm}
\caption{Parameter search for $\epsilon$ when fixing $\alpha=4.0$.} 
\label{table:search_conf}
\end{minipage}
~~~~~~~
\begin{minipage}{\columnwidth}
\begin{center}
\footnotesize
\begin{tabular}{L{0.65cm} |X{1.33cm} |X{1.33cm} X{1.33cm} X{1.33cm}}\hline
$\alpha$ & All & Many & Medium & Few \\\toprule
\!1.0\!\!       & 80.3 & 84.8 & 81.2 & 63.2\\ 
\!2.0\!\!       & 81.1 & \textbf{85.6} & 82.3 & 60.3 \\
\!4.0\!\!       & \textbf{81.6} & 85.5 & \textbf{82.6} & \textbf{63.1} \\
\!8.0\!\!       & 81.3 & 85.1 & 82.5 & 60.1 \\ \bottomrule
\end{tabular}
\end{center}
\vspace*{-0.25cm}
\caption{Parameter search for $\alpha$ when fixing $\epsilon=0.975$.} 
\label{table:search_alpha}
\end{minipage}
\vspace*{-0.25cm}
\end{table*}

\begin{table*}[t!]
\begin{center}
\scriptsize
\begin{tabular}{L{1.6cm} |X{1.3cm} |X{0.97cm} X{0.97cm} X{0.97cm} X{0.97cm} X{0.97cm} |X{0.97cm} X{0.97cm} X{0.97cm} X{0.97cm} X{0.97cm}}\hline
\multicolumn{2}{c|}{Class Group} & \multicolumn{5}{c|}{MS-COCO} & \multicolumn{5}{c}{Pascal-VOC} \\\toprule
Category & Method & Mis. 20\% & Mis. 40\% & \!\!Rand. 20\%\!\! & \!\!Rand. 40\%\!\! & Single & Mis. 20\% & Mis. 40\% & \!\!Rand. 20\%\!\! & \!\!Rand. 40\%\!\! & Single \\\midrule
{Default} 
& BCE Loss & 73.1$\pm$0.4 & 63.8$\pm$0.8 & 59.8$\pm$0.5 & 43.5$\pm$0.8 & 69.7$\pm$1.8 & 82.9$\pm$0.1 & 75.4$\pm$0.3 & 76.3$\pm$0.5 & 72.0$\pm$5.5 & 85.7$\pm$0.1 \\ \midrule
\multirow{2}{*}{Noisy Labels} 
& Co-teaching & 82.5$\pm$0.1 & 78.6$\pm$0.4 & 65.5$\pm$0.0 & 43.6$\pm$0.1 & 68.1$\pm$1.4 & 92.5$\pm$0.3 & 90.9$\pm$0.1 & 82.6$\pm$0.4 & 70.0$\pm$0.4 & 81.9$\pm$1.6   \\ 
& ASL & 82.8$\pm$0.1 & 80.3$\pm$0.2 & 75.0$\pm$0.1 & 66.2$\pm$0.1 & 73.3$\pm$0.2 & 91.2$\pm$0.0 & 86.4$\pm$0.1 & 90.1$\pm$0.1 & 74.8$\pm$0.3 & 86.8$\pm$0.1  \\ \midrule
\multirow{2}{*}{Missing Labels} 
& LL-R & 80.7$\pm$0.1 & 75.3$\pm$0.2 & 74.0$\pm$0.2 & 69.3$\pm$0.2 & 74.2$\pm$0.2 & 87.5$\pm$0.5 & 83.1$\pm$1.2 & 85.5$\pm$0.3 & 78.6$\pm$2.2 & 89.1$\pm$0.4    \\
& LL-Ct & 81.3$\pm$0.0 & 77.4$\pm$0.1 & 73.2$\pm$0.3 & 70.1$\pm$0.1 & 76.9$\pm$0.1 & 88.8$\pm$0.5 & 84.8$\pm$1.4 & 87.1$\pm$0.2 & 78.8$\pm$0.3 & 89.3$\pm$0.1   \\ \midrule
Proposed & BalanceMix & 84.3$\pm$0.1 & 81.6$\pm$0.3 & 76.5$\pm$0.1 & 74.5$\pm$0.1 & 77.4$\pm$0.1 & 92.9$\pm$0.1 & 92.0$\pm$0.0& 91.2$\pm$0.1 & 84.4$\pm$0.1 & 92.6$\pm$0.1 \\ \bottomrule
\end{tabular}
\end{center}
\vspace*{-0.25cm}
\caption{Last mAPs on MS-COCO and Pascal-VOC with standard errors.} 
\label{table:std_error}
\vspace*{-0.5cm}
\end{table*}

\begin{table*}[t]
\begin{center}
\footnotesize
\begin{tabular}{L{2.0cm} |L{1.6cm} |X{1.02cm} X{1.02cm} X{1.02cm} |X{1.02cm} X{1.02cm} X{1.02cm} |X{1.02cm} X{1.02cm} X{1.02cm} }\hline
\multicolumn{2}{c|}{Class Group} & \multicolumn{3}{c|}{All} & \multicolumn{3}{c|}{Medium-shot} & \multicolumn{3}{c}{Few-shot} \\\toprule
Category & Method & 0\% & 20\% & 40\% & 0\% & 20\% & 40\% & 0\% & 20\% & 40\%  \\\midrule
{Default} 
& BCE Loss & 87.7 & 82.9 & 75.4 & 94.1 & 85.3 & 77.1 & 84.9 & 81.8 & 74.8 \\ \midrule
\multirow{3}{*}{Noisy Labels} 
& Co-teaching \!\!\!\! & 90.8 & \underline{92.5} & \underline{90.9} & \underline{93.9}&  \underline{94.0}& \underline{92.8} & 89.4 & \underline{91.8} & \underline{90.0} \\ 
& ASL & \underline{91.4} & 91.2 & 86.4 & 92.9&  92.4& 80.5 & \underline{90.8}  &90.7  &88.9 \\ 
& T-estimator\!\!\!\! & 91.0 & 89.5 & 89.0 & 92.4 & 91.9 & 91.6 & 90.2 & 88.5 & 87.9 \\ \midrule
\multirow{2}{*}{Missing Labels} 
& LL-R  & 81.8 & 87.5 & 83.1 & 92.5 & 89.6 & 85.7 & 77.2 &86.6  &81.9  \\
& LL-Ct  & 84.0& 88.8 & 84.8 &90.3  & 90.7&  87.9& 81.3 & 87.9 & 83.5 \\ \midrule
Proposed & BalanceMix &\textbf{93.3} & \textbf{92.9} &\textbf{92.0}  &\textbf{95.1}  &\textbf{94.4} & \textbf{93.6} &\textbf{92.5} & \textbf{92.2} & \textbf{91.3}\\ \bottomrule
\end{tabular}
\end{center}
\vspace*{-0.4cm}
\caption{Last mAPs on Pascal-VOC with \textbf{mislabeling} of $0$--$40\%$. The 1st and 2nd best values are in bold and underlined.} 
\label{table:voc_mislabeling}
\vspace*{-0.1cm}
\end{table*}

\begin{table*}[t]
\begin{center}
\footnotesize
\begin{tabular}{L{2.0cm} |L{1.6cm} |X{1.02cm} X{1.02cm} X{1.02cm} |X{1.02cm} X{1.02cm} X{1.02cm} |X{1.02cm} X{1.02cm} X{1.02cm} }\hline
\multicolumn{2}{c|}{Class Group} & \multicolumn{3}{c|}{All} & \multicolumn{3}{c|}{Medium-shot} & \multicolumn{3}{c}{Few-shot} \\\toprule
Category & Method & 0\% & 20\% & 40\% & 0\% & 20\% & 40\% & 0\% & 20\% & 40\%  \\\midrule
{Default} 
& BCE Loss & 87.7 & 76.3 &  72.0&94.1 &80.0  & 74.4 & 84.9 & 74.8 & 71.0\\ \midrule
\multirow{3}{*}{Noisy Labels} 
& Co-teaching \!\!\!\! & 90.8  &82.6  & 70.3 & \underline{93.9}&  87.1& 82.8 & 89.4 & 80.6 &64.9 \\ 
& ASL & \underline{91.4} & \underline{90.1} & \underline{74.8} & 92.9& \underline{92.3} & \underline{87.6} & \underline{90.8} & \underline{89.2} &69.3 \\ 
& T-estimator \!\!\!\! & 91.0 & 85.9 & 70.1 & 92.4 & 89.3 & 80.3 & 90.2 & 84.4 & 65.6 \\ \midrule
\multirow{2}{*}{Missing Labels} 
& LL-R  & 81.8 & 85.5 & 78.6 &92.5 & 88.8 & 83.0 & 77.2 & 84.0 & 76.8 \\
& LL-Ct & 84.0& 87.1 &78.8  & 90.3 &89.3 &82.9  & 81.3 &86.1  & \underline{77.1} \\ \midrule
Proposed & BalanceMix & \textbf{93.3} & \textbf{91.2}  & \textbf{84.4} & \textbf{95.1}& \textbf{93.8} & \textbf{91.7}  & \textbf{92.5} & \textbf{90.0} & \textbf{81.3} \\ \bottomrule
\end{tabular}
\end{center}
\vspace*{-0.4cm}
\caption{Last mAPs on Pascal-VOC with \textbf{random flipping} of $0$--$40\%$. The 1st and 2nd best values are in bold and underlined.} 
\vspace*{-0.35cm}
\label{table:voc_random_flip}
\end{table*}

\section{D. Detailed Experiment Configuration}

All the algorithms are implemented using Pytorch 21.11 and run using two NVIDIA V100 GPUs utilizing distributed data parallelism.
We fine-tune ResNet-50 pre-trained on ImageNet-1K for 20, 50, and 40 epochs for Pascal-VOC (a batch size of $32$), MS-COCO (a batch size of $64$), and DeepFashion (a batch size of $64$) using an SGD optimizer with a momentum of 0.9 and a weight decay of $10^{-4}$. All the images are resized with $448\times448$ resolution. The initial learning rate is set to be $0.01$ and decayed with a cosine annealing without restart. The number of  warm-up epochs is set to be 5, 10, and 5 for the three datasets, respectively. We adopt a state-of-the-art Transformer-based decoder\,\cite{lanchantin2021general, liu2021query2label, ridnik2023ml} for the classification head. These experiment setups are exactly the same for all compared methods.

The hyperparameters for the compared methods are configured favorably, as suggested in the original papers.
\begin{itemize}[leftmargin=10pt]
\setlength\itemsep{0em}
\item {Co-teaching}\,\cite{han2018co}: We extend the vanilla version to support multi-label classification. Two models are maintained for co-training. Instead of using the known noise ratio, we fit a bi-modal univariate GMM to the losses of all instances, i.e., instance-level modeling. Then, the instances whose probability of being clean is greater than $0.5$ are selected as clean instances.
\item {ASL}\,\cite{ben2020asymmetric}: Three hyperparameters -- $\gamma^{+}$ which is a down-weighting coefficient for positive labels, $\gamma^{-}$ which is a down-weighting coefficient for negative labels, and $m$ which is a probability margin -- are set to be $0.0$, $4.0$, and $0.05$, respectively.
\item {LL-R \& LL-Ct}\,\cite{kim2022large}: The only hyperparameter is $\Delta_{rel}$, which determines the speed of increasing the rejection (or correction) rate. The default value used in the original paper was $0.2$ for 10 epochs. Hence, we modify the value according to our training epochs, such that the rejection (or correction) ratios at the final epoch are the same. Specifically, it is set to be $0.1$ for $20$ epochs (Pascal-VOC), $0.04$ for $50$ epochs (MS-COCO), and $0.05$ for $40$ epochs (DeepFashion), respectively.
\end{itemize} %The three compared methods use RandAug and Mixup for data augmentation, while the default method (BCE) only uses RandAug; we contrast the improvement of using Mixup with BCE in Table \ref{table:com_analysis}.

Regarding the state-of-the-art comparison with ResNet-101 and TResNet-L, we follow exactly the same settings in the backbone, image resolution, and data augmentation \cite{ridnik2023ml}. See Table 5 for details.

\section{E. Hyperparameters}
\algname{} introduces two hyperparameters: $\epsilon$, a confidence threshold for re-labeling and $\alpha$, the parameter of the beta distribution for Mixup. We search for a suitable pair of these two hyperparameters based on MS-COCO.

First, we fix $\alpha\!=\!4.0$ and conduct a grid search to find the best $\epsilon$, as summarized in Table \ref{table:search_conf}. Intuitively, a high threshold value achieves high precision in re-labeling, while a low threshold value achieves high recall. For mislabeling, high precision is more beneficial than high recall; thus, the interval of $0.950$--$0.975$ exhibits the best mAP. However, in the missing\,(single positive) label setup, high recall precedes high precision because increasing the amount of positive labels is more beneficial; thus, the interval of $0.550$--$0.600$ exhibits the best mAP. Overall, we use $\epsilon=0.550$ for the missing label setup, while $\epsilon=0.975$ for other setups. 

Second, we fix $\epsilon=0.975$ and repeat a grid search for the best $\alpha$. Table \ref{table:search_alpha} summarizes the mAPs on MS-COCO with a mislabeling ratio of $40\%$. The best mAP for many-shot classes is observed when $\alpha=2.0$. However, the overall mAP of \algname{} is the best when $\alpha=4.0$ owing to the highest mAP on medium-shot and few-shot classes. Therefore, we use $\alpha=4.0$ in general.

These hyperparameter values found may not be optimal as we validate them only in a few experiment settings, but \algname{} shows satisfactory performance with them in all the experiments presented in the paper. We believe that the performance of \algname{} could be further improved via a more sophisticated parameter search. % We leave this as future work. % Future work 치고는 너무 별볼일 없음

\section{F. Additional Main Results}

\subsection{Results with Standard Errors}
Table \ref{table:std_error} summarizes the last mAPs on MS-COCO and Pascal-VOC. We repeat the experiments thrice and report the averaged mAPs as well as their standard errors. These standard errors are, in general, very small.

\begin{table}[t!]
\begin{center}
\footnotesize
\begin{tabular}{L{1.6cm} |X{0.8cm} X{0.8cm} X{0.8cm} X{0.8cm} X{0.8cm}}\hline
Mixing coef. $\alpha$ & 0.0 & 1.0 & 2.0 & 4.0 & 8.0 \\\toprule
%All  & 78.1 & 80.3 & 81.3 & 81.6 & 81.1 \\\midrule
Many-shot & 85.4 & 85.6 & 85.1 & 85.5 & 84.8 \\
Few-shot  & 57.4 & 60.3 & 62.4 & 63.1 & 63.2 \\ \bottomrule
\end{tabular}
\end{center}
\vspace*{-0.4cm}
\caption{Varying $\alpha$ on COCO with mislabeling of $40\%$.} 
\label{table:alpha}
\vspace*{-0.0cm}
\end{table}

\subsection{Results on Pascal-VOC}
Tables \ref{table:voc_mislabeling} and \ref{table:voc_random_flip} summarize the mAPs on Pascal-VOC with mislabeing and random flipping. The performance trends are similar to those on MS-COCO except that Co-teaching exhibits higher mAPs than ASL in the mislabeling noise. In Pascal-VOC unlike MS-COCO, the number of positive labels per instance is only two on average. Therefore, the instance-level selection of Co-teaching can perform better than ASL. However, in the random flipping noise where even negative labels are flipped by a given noise ratio, Co-teaching is much worse than ASL. \algname{} consistently exhibits the best mAPs for all class categories. Regarding T-estimator, it performs much better than BCE Loss and exhibits comparable performance to Co-teaching and ASL, even if 10\% of training data is not used for training since it is required for the noisy validation set.

% , which means that it is a problem closer to single-label classification. JGL: 어찌되었던 label이 2개인데 괜히 트집잡힐 말임

\section{G. Analysis of Label-wise Management}

\begin{table}[t]
\begin{center}
\footnotesize
\begin{tabular}{L{2.4cm} |X{0.75cm} X{1.0cm} X{1.3cm} X{1.1cm}}\hline
Method & Clean & \!\!\!Mis. 40\%\!\!\! & \!\!\!Rand. 40\%\!\!\! & Missing \!\!\\\toprule
\!\!\!\!Co-teaching  & 82.8 & 78.6 & 43.6 & 68.1 \\
\!\!\!\!T-estimator & 84.3 & 80.5 & 69.9 & 16.8 \\
\!\!\!\!LL-R  & 82.5 & 75.3 & 69.3 & 74.2 \\
\!\!\!\!LL-Ct & 79.4 & 77.4 & 70.1 & 76.9 \\
\!\!\!\!BalanceMix\,(wo Min.)\!\!\!\!\!\!\! & \textbf{85.0} & \textbf{80.9} & \textbf{73.0} & \textbf{77.0} \\ \bottomrule
\end{tabular}
\end{center}
\vspace*{-0.4cm}
\caption{Analysis of \algname{} {w.o} using the minority sampler on MS-COCO.} 
\label{table:refine}
\vspace*{-0.0cm}
\end{table}

 \begin{table*}[t!]
\begin{center}
\footnotesize
\begin{tabular}{L{2.3cm} |X{1.3cm} X{1.3cm} |X{1.3cm} X{1.3cm} ||X{1.3cm} X{1.5cm} |X{1.4cm} X{1.3cm} }\hline
& \multicolumn{4}{c||}{Clean Label Selection (C by Eq.\,\eqref{eq:clean_p})} &  \multicolumn{4}{c}{Re-labeling (R by Eq.\,\eqref{eq:relabeling})} \\\toprule
Noise Type\!\!           & \multicolumn{2}{c|}{Mislabel 20\%} & \multicolumn{2}{c||}{Mislabel 40\%}  & \multicolumn{2}{c|}{Mislabel 20\%} & \multicolumn{2}{c}{Mislabel 40\%}\\\midrule
Training Progress\!\!\!\!   & Precision & Recall & Precision & Recall  & \!\!Proportion\!\! & Accuracy & Proportion & Accuracy \\\midrule
25\% Epochs    & 99.2\% & 85.3\% & 96.1\% & 90.5\% & 10.1\% & 98.6\% & 12.0\% & 98.9\%\\ 
50\% Epochs    & 99.0\% & 88.9\% & 95.5\% & 92.7\% & 9.1\% & 98.6\% & 11.2\% & 98.8\%\\
%75\% Epochs    & 98.8\% & 90.8\% & 94.7\% & 94.1\% & 8.6\% & 98.6\% & 9.5\%  & 98.6\%\\
100\% Epochs   & 98.6\% & 91.5\% & 94.5\% & 94.3\% & 8.3\% & 98.5\% & 9.1\%  & 98.5\%\\ \bottomrule
\end{tabular}
\end{center}
\vspace*{-0.35cm}
\caption{Clean label selection and re-labeling performance of \algname{} on MS-COCO: 2nd-5th columns summarize the label precision and label recall of selecting clean labels, and 6th-9th columns summarize the proportion of re-labeled labels and their re-labeling accuracy.} 
\label{table:precision_recall}
\vspace*{-0.1cm}
\end{table*}

%\begin{comment}
\begin{table*}[t!]
\begin{center}
\footnotesize
\begin{tabular}{L{2.3cm} |X{1.3cm} X{1.3cm} |X{1.3cm} X{1.3cm} ||X{1.3cm} X{1.5cm} |X{1.4cm} X{1.3cm} }\hline
& \multicolumn{4}{c||}{Clean Label Selection (C by Eq.\,(8))} &  \multicolumn{4}{c}{Re-labeling (R by Eq.\,(9))} \\\toprule
Noise Type\!\!           & \multicolumn{2}{c|}{Mislabel 20\%} & \multicolumn{2}{c||}{Mislabel 40\%}  & \multicolumn{2}{c|}{Mislabel 20\%} & \multicolumn{2}{c}{Mislabel 40\%}\\\midrule
Training Progress\!\!\!\!   & Precision & Recall & Precision & Recall  & \!\!Proportion\!\! & Accuracy & Proportion & Accuracy \\\midrule
25\% Epochs    & 99.6\% & 88.0\% & 98.2\% & 91.5\% & 2.9\% & 97.4\% & 2.0\% & 97.4\%\\ 
50\% Epochs    & 99.3\% & 93.6\% & 97.2\% & 94.8\% & 6.9\% & 98.8\% & 5.9\% & 98.8\%\\ 
75\% Epochs    & 99.2\% & 94.4\% & 97.0\% & 95.5\% & 8.6\% & 99.0\% & 7.0\% & 99.0\%\\ 
100\% Epochs   & 99.2\% & 95.0\% & 96.7\% & 95.9\% & 8.6\% & 99.1\% & 6.8\% & 99.1\%\\ \bottomrule
\end{tabular}
\end{center}
\vspace*{-0.35cm}
\caption{Clean label selection and re-labeling performance of \algname{} on Pascal-VOC: 2nd-5th columns summarize the label precision and label recall of selecting clean labels, and 6th-9th columns summarize the proportion of re-labeled labels and their re-labeling accuracy.} 
\label{table:relabel_performance}
\vspace*{-0.45cm}
\end{table*}
%\end{comment}

\subsection{G.1. Mixing with Different Diversity} 

The diversity is added by mixing the instances from the random sampler with the instances from the minority sampler via Mixup. Thus, when the Mixup coefficient $\alpha$ is $0$, mixing is not performed at all, and the diversity is the lowest. On the other hand, as $\alpha$ becomes larger, minority samples are more strongly mixed with random samples, and the diversity gets higher. As shown in Table \ref{table:alpha}, increasing the value of $\alpha$ enhances few-shot class performance, but excessive adjustments degrade many-shot class performance; when $\alpha=0$, a low performance of few-shot classes is attributed to the overfitting caused by limited diversity.

\subsection{G.2. Pure Effect of Label-wise Management} 
% \subsection{Fair Comparison on Label Refinement} 

We compare solely the label-wise management with the refinement-based methods (Co-teaching, T-estimator, LL-R, and LL-Ct) by excluding the additional gains from the minority sampler. Thus, we replace the minority sampler of \algname{} with the random sampler because the refinement-based methods use the random Mixup. Then, the remaining differences of BalanceMix from others are (1) the definition of ambiguous labels and (2) the diminution of their loss based on inferred clean probabilities. Table \ref{table:refine} shows the result of such comparison, clearly showing the \emph{pure} superiority of our label-wise refinement over other counterparts.
%

% Label Precision and Label Recall.
\subsection{G.3. Label Precision and Label Recall} 

% Proportion of Re-labeling and Accuracy.
The label-wise management of \algname{} involves selecting clean labels and re-labeling incorrect labels. There are four metrics to evaluate clean label selection and re-labeling performance. Regarding label selection, there are two indicators, label precision and recall, of evaluating how accurate and how many clean labels are chosen from noisy labels, respectively\,\cite{han2018co, song2022learning}. For convenience, let ${\sf C}$ be the set of all selected labels from all noisy labels, and ${\sf L}$ be the set of all clean labels. Then, the label precision and recall are formulated as:
\begin{equation}
\begin{gathered}
{\rm Label~Precision}={|\{\tilde{y} \in {\sf C} : \tilde{y} = y\}|}~/~{|{\sf C}|},\\
{\rm Label~Recall}={|\{\tilde{y} \in {\sf C} : \tilde{y} = y\}|}~/~{|{\sf L}|}.
\end{gathered}
\end{equation}
Regarding re-labeling, we evaluate its performance based on the proportion of re-labeled labels and their re-labeling accuracy. Let ${\sf R}$ be the re-labeled labels among all noisy labels, and ${\sf D}$ be the entire label in data. Then, the proportion and accuracy are formulated as:
\begin{equation}
\begin{gathered}
{\rm Relabel~Proportion}={|{\sf R}|}~/~{|{\sf D}|},\\
{\rm Relabel~Accuracy}={|\{\tilde{y} \in {\sf R} : \tilde{y} = y\}|}~/~{|{\sf R}|}.
\end{gathered}
\end{equation}

%

%\smallskip\smallskip
%\noindent\textbf{Label Selection and Re-labeling.} 

Table \ref{table:precision_recall} summarizes their performance on MS-COCO at three different learning progress. 
%We provide additional results for the label-wise management including a comparison without using the minority sampler in Appendix G.
%We provide additional results for Pascal-VOC including detailed definitions of used performance metrics in Appendix G.
For label selection, we evaluate label precision and recall\,\cite{han2018co, song2022learning} of the selected clean labels, where they are indicators of how accurate and how many clean labels are chosen, respectively. \algname{} exhibits very high precision and recall, and the recall increases greatly as training progresses without compromising the precision. For re-labeling, we evaluate the percentage of re-labeled labels and their accuracy. \algname{} keeps very high re-labeling accuracy in all training phases. Thus, as the model continues to evolve, more clean labels are selected with high precision, and incorrect labels are re-labeled with high accuracy. 

Table \ref{table:relabel_performance} summarizes their performance on Pascal-VOC at three different learning progress. In Pascal-VOC, \algname{} exhibits similar trends of label selection, compared when using MS-COCO. However, we observe that the number of re-labeled labels shows a different trend in MS-COCO and Pascal-VOC. The number increases over training epochs in Pascal-VOC, but an opposite trend is observed in MS-COCO. We expect that the re-labeling performance may be associated with the learning difficulty of training data and the number of classes in training data. We will leave further analysis of re-labeling as future work.

\begin{figure*}[t!]
\begin{center}
\includegraphics[width=17.2cm]{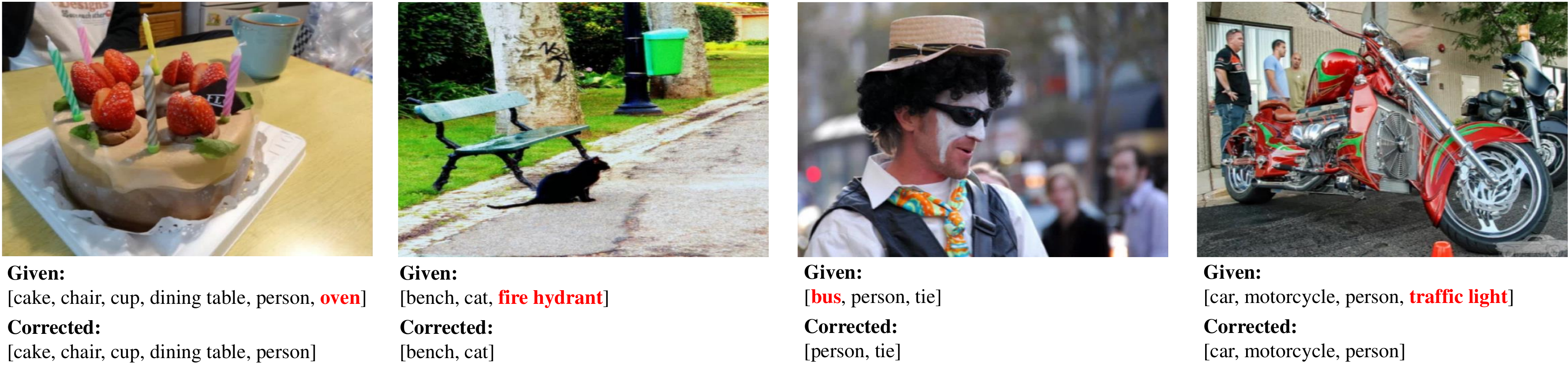}
\end{center}
\begin{center}
\includegraphics[width=17.2cm]{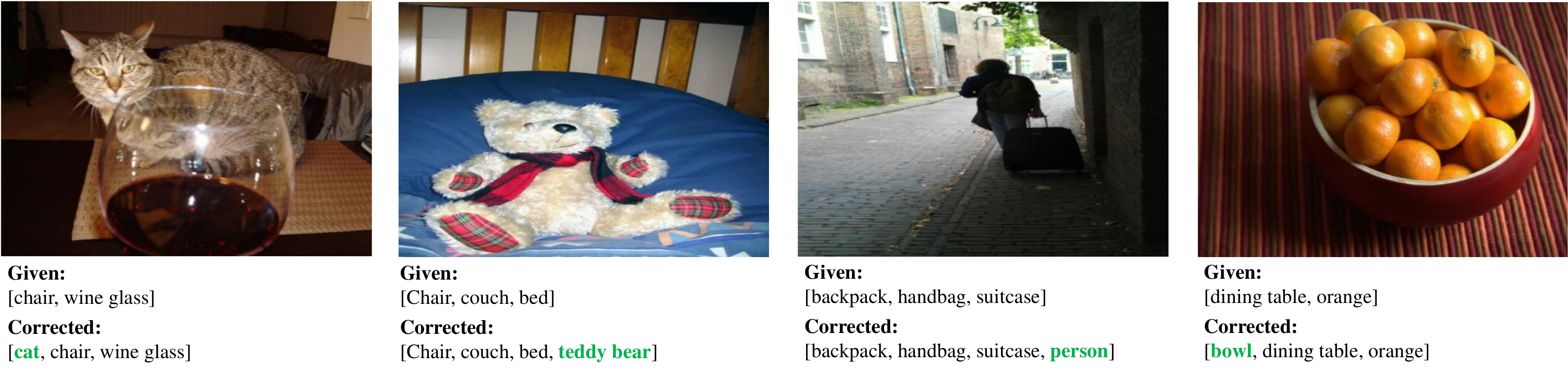}
\end{center}
\vspace*{-0.35cm}
\caption{Examples of incorrect and missing labels originally contained in MS-COCO. The {\color{red}red} labels (incorrect labels) and {\color{green}green} labels (missing labels) are detected and corrected by \algname{} during training.}
\label{fig:examples}
\vspace*{-0.4cm}
\end{figure*}

\begin{comment}
\begin{table}[t!]
\begin{center}
\footnotesize
\begin{tabular}{L{1.65cm} |X{1.12cm} X{1.12cm} || X{1.12cm} X{1.12cm} }\hline
& \multicolumn{2}{c||}{Label Selection} &  \multicolumn{2}{c}{Re-labeling} \\\toprule
Noise Type\!\!          & \multicolumn{2}{c||}{Mislabel 40\%}  & \multicolumn{2}{c}{Mislabel 40\%}\\\midrule
Train Progress\!\!\!\!   & Precision\!\!\! & Recall & Proportion\!\!\! & Accuracy \\\midrule
25\% Epochs    & 96.1\% & 90.5\% & 12.0\% & 98.9\%\\ 
50\% Epochs    & 95.5\% & 92.7\% & 11.2\% & 98.8\%\\
100\% Epochs   & 94.5\% & 94.3\% & 9.1\%  & 98.5\%\\ \bottomrule
\end{tabular}
\end{center}
\vspace*{-0.25cm}
\caption{Clean label selection and re-labeling performance of \algname{} on MS-COCO: 2nd-3th columns summarize the label precision and label recall of selecting clean labels, and 4th-5th columns summarize the proportion of re-labeled labels and their re-labeling accuracy. See Appendix G for the details of used metrics.} 
\label{table:precision_recall}
\vspace*{-0.4cm}
\end{table}
\end{comment}

% https://github.com/meetps/pytorch-semseg/blob/801fb200547caa5b0d91b8dde56b837da029f746/ptsemseg/loader/pascal_voc_loader.py#L75

\section{H. Noisy Labels in MS-COCO}

It is of interest to see a significant improvement of \algname{} on MS-COCO in our state-of-the-art comparison (see Table 5). It turns out that MS-COCO originally has incorrect and missing labels. Fig.~\ref{fig:examples} shows a few successfully re-labeled examples from MS-COCO by \algname{}. The first row shows four examples with incorrect labels, and the second row shows four examples with missing labels. As an example with the first image, an oven is mislabeled as a positive label, but it is re-labeled as a negative one by \algname{}. As an example with the fifth image, a cat is omitted in labeling, but it is re-labeled as a positive one by \algname{}. Therefore, the state-of-the-art performance of \algname{} in MS-COCO is attributed to its versatility for real noisy and imbalanced labels.

\begin{comment}
\begin{table}[t]
\vspace*{-0.0cm}
\begin{center}
\footnotesize
\begin{tabular}{L{2cm} |X{0.8cm} X{1.05cm} X{1.35cm} X{0.95cm}}\hline
Method & Clean & \!\!\!Mis. 40\%\!\!\! & \!\!\!Rand. 40\%\!\!\! & Missing \\\toprule
\!\!\!\!Co-teaching  & 82.8 & 78.6 & 43.6 & 68.1 \\
\!\!\!\!T-estimator & 84.3 & 80.5 & 69.9 & 16.8 \\
\!\!\!\!LL-R  & 82.5 & 75.3 & 69.3 & 74.2 \\
\!\!\!\!LL-Ct & 79.4 & 77.4 & 70.1 & 76.9 \\
\!\!\!\!BalanceMix\,(wo Min.)\!\!\!\! & \textbf{85.0} & \textbf{80.9} & \textbf{73.0} & \textbf{77.0} \\ \bottomrule
\end{tabular}
\end{center}
\vspace*{-0.25cm}
\caption{Comparison \textbf{w.o} using minority sampler on MS-COCO.} 
\label{table:refine}
\vspace*{-0.35cm}
\end{table}

\smallskip\smallskip
\noindent\textbf{Comparison with Refinement-based Methods.} We make a more fair comparison with the refinement-based methods (i.e., Co-teaching, T-estimator, LL-R, and LL-Ct) by excluding the additional gains from the minority sampler. Thus, we replace the minority sampler of \algname{} with the random sampler because the refinement-based methods use random Mixup. Table \ref{table:refine} shows the result of such \textrm{``fair'' comparison}, clearly showing the superiority  of our label-wise refinement. The major difference of BalanceMix from others is (1) the definition of ambiguous labels and (2) the down-weight of their loss based on inferred clean probabilities.

\end{comment}